\pdfoutput=1
\documentclass[11pt]{article}

\usepackage{ACL2023}
\usepackage{times}
\usepackage{latexsym}

\usepackage{xcolor}
\usepackage{colortbl}
\usepackage{booktabs}
\usepackage{graphicx}
\usepackage{algorithmic}
\usepackage{algorithm}
\usepackage{amsmath}
\usepackage{adjustbox}
\usepackage{lipsum}
\usepackage{float}
\usepackage{multicol}
\usepackage{multirow}
\usepackage{stfloats}
\usepackage{multirow}
\usepackage{array}
\usepackage[T1]{fontenc}
\usepackage[utf8]{inputenc}

\usepackage{microtype}

\usepackage{inconsolata}

\newcommand\methodname{\textcolor{black}{\textsc{LinguAlchemy}}}

\definecolor{linecolor}{HTML}{4C5760}
\definecolor{shadecolor}{HTML}{6C6C6C}

\definecolor{darkgreen}{rgb}{0.0, 0.5, 0.0}
\definecolor{orange}{rgb}{1.0, 0.65, 0.0}
\definecolor{darkred}{rgb}{0.55, 0.0, 0.0}

\title{$\vcenter{\hbox{\includegraphics[scale=0.035]{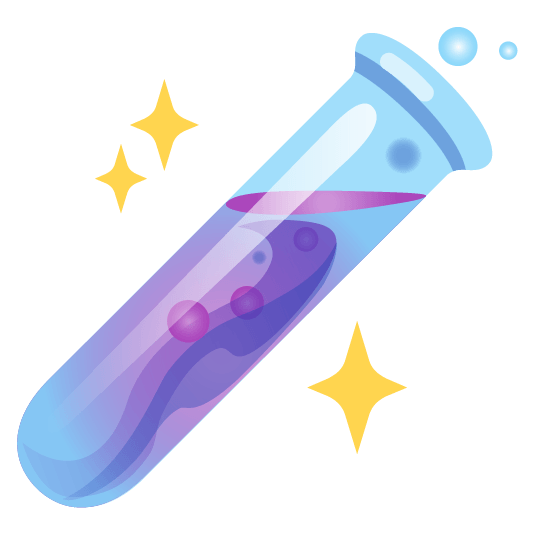}}}$LinguAlchemy: Fusing Typological and Geographical Elements\\ for Unseen Language Generalization}
\author{Muhammad Farid Adilazuarda$^1$, Samuel Cahyawijaya$^3$, \textbf{Genta Indra Winata$^4$\thanks{*The work was done prior joining Capital One}}, \\ \textbf{Ayu Purwarianti}$^2$, \textbf{Alham Fikri Aji}$^1$ \\
  $^1$MBZUAI \quad $^2$Institut Teknologi Bandung  \\
   $^3$The Hong Kong University of Science and Technology  \quad $^4$Capital One\\
  \texttt{farid.adilazuarda@mbzuai.ac.ae}}

\begin{document}
\maketitle
\begin{abstract}

Pretrained language models (PLMs) have become remarkably adept at task and language generalization. Nonetheless, they often fail when faced with unseen languages. In this work, we present \methodname, a regularization method that incorporates various linguistic information covering typological, geographical, and phylogenetic features to align PLMs representation to the corresponding linguistic information on each language. \methodname~significantly improves the performance of mBERT and XLM-R on low-resource languages in multiple downstream tasks such as intent classification, news classification, and semantic relatedness compared to standard approach and displaying a high degree of unseen language generalization. We further introduce \textsc{AlchemyScale} and \textsc{AlchemyTune}, extension of \methodname~which adjusts the linguistic regularization weights automatically, alleviating the need for hyperparameter search. 
\end{abstract}

\section{Introduction}
% P1: Large Language Model Is Powerful but it has a poor generalization towards unseen languages. asai2023buffer, belebele. zhang2023 also shows similar issues on code-mixing data which can be also considered as a case of unseen languages.

Significant advancements in language processing technology have been achieved through the development of PLMs with their impressive capability in language comprehension and generation~\cite{devlin-etal-2019-bert, liu2019roberta, lewis2019bart,li2021pretrained, sanh2022multitask,raffel2023exploring}. The development has been further expanded to non-English languages~\cite{conneau2020unsupervised, martin2020camembert, wilie-etal-2020-indonlu,kakwani2020indicnlpsuite,cahyawijaya2024cendol}. However, there is still a gap in these models' ability to generalize effectively to low-resource and unseen languages, although there have been a numerous work in the field~\cite{pfeiffer-etal-2021-unks, goyal2021largerscale,alabi2022adapting, ebrahimi2022americasnli,yong2023bloom+}.

% P2: Existing method on unseen language generalization
% https://aclanthology.org/2023.emnlp-main.431/, https://aclanthology.org/2021.findings-emnlp.63.pdf
% https://aclanthology.org/2020.emnlp-main.180.pdf, https://aclanthology.org/2021.findings-emnlp.410.pdf
% Answer: [trim={left bottom right top},clip]

\begin{figure}[t!]
    \centerline{\includegraphics[width=1\columnwidth]{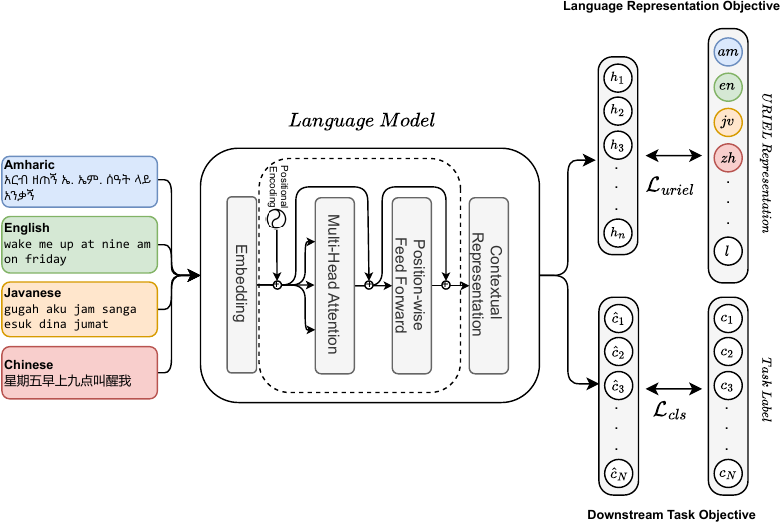}}
    \caption{\textsc{LinguAlchemy} enhances performance in unseen languages by allowing the model to predict the linguistic vector and then fitting it via a similarity loss towards the specific language's URIEL vector.}
    \label{fig:lingualchemy-hook}
\end{figure}

Previous approaches \cite{rathore-etal-2023-zgul, ustun-etal-2022-udapter, pfeiffer-etal-2020-mad, ansell-etal-2021-mad-g}  are based on the assumption that disregards the fact that in a real-world application, there is usually no language information from the user, highlighting the importance of the multilingual robustness of a language model. The second assumption might cause performance degradation due to the error propagation from the language identification module~\cite{adilazuarda2023obscure}. However, these methods inherit the limitations of the pretrained multilingual models, such as the limited capacity to adapt effectively to low-resource and unseen languages. Furthermore, while the framework facilitates adaptation to specific target languages, it may bias the model towards them, potentially impacting its performance on other languages.

In this work, we introduce \methodname{}, a novel method that incorporates a unified representation across multiple languages to enable the model for utilizing the shared linguistic knowledge. Our approach differs from adapter-based approaches which often segment language understanding into multiple, isolated language-specific modules. Instead, we developed a regularization technique that utilizes linguistic information directly into the model’s architecture, allowing for language-agnostic inference. Our evaluations demonstrate that \methodname{} not only enhances generalization capabilities of mBERT~\cite{devlin2018bert} and XLM-R~\cite{conneau2020unsupervised} on unseen languages but also upholds robust performance across high-resource languages, all without prior knowledge of the query's language.

Our strategy aims to refine cross-lingual generalization by leveraging linguistic features encapsulated in URIEL vectors. We hypothesize that languages with similar syntactic and geographical characteristics can benefit from shared representational frameworks, significantly boosting performance in multilingual settings. This approach is particularly beneficial in contexts where language resources are limited.

% P4: Contribution List

In summary, our contributions are as follows:
\begin{enumerate}
    \item We propose \methodname{}, a regularization method that utilizes geographical and syntactic information to foster the models' unified representation. 
    \item \methodname{} does not require any architectural change and can be adapted to different tasks and models.
    \item We demonstrate strong performance on 50+ languages across three diverse datasets and tasks (intent classification, news classification, and semantic relatedness) for models trained with \methodname{}, including the languages that are not seen during pretraining.
    \item We introduce two automatic hyperparameter search methods to scale the classification and auxiliary loss factors used in the fine-tuning stage, namely dynamiclearn and dynamicscale.
\end{enumerate}

\section{Related Work}
PLMs with their transformer-based architectures have been demonstrating exceptional capabilities in language comprehension and generation \citet{ganesh2021transformers}. \citet{rathore-etal-2023-zgul}~have explored how these models learn intricate linguistic features, including syntax and semantics to enhance their performance across a wide range of language tasks.

% Research in this area 

Incorporating new unseen languages has been a longstanding problem in the multilingual research, MAD-X~\cite{pfeiffer-etal-2020-mad} employ a language adapter to learn new unseen languages using language adapters that mitigate the risk of forgetting pre-trained knowledge, which is known as the \textit{curse-of-multilinguality} \cite{conneau2020unsupervised}. Nonetheless, this approach requires training for generalizing to new unseen languages, which makes it costly and difficult to scale to thousands of languages. MAD-G~\cite{ansell-etal-2021-mad-g} and Udapter~\cite{ustun-etal-2020-udapter} further generalize this approach by utilizing a linguistic-driven contextual parameter generator (CPG) module to generate language-specific parameters, allowing the models to generalize to other languages with similar linguistic characteristics. Recently, \citet{rathore-etal-2023-zgul} introduce ZGUL, which combines representations over multiple language adapters and adding linguistic vector information to generate the unseen language representation. Despite the effectiveness, all these approaches rely on two assumptions: (1) strict categorization of languages and (2) knowing the language category of the query apriori—our definition of ``a priori categorization" as incorporating language-specific information into the model.

In parallel, the development of linguistically-driven resources such as the URIEL vector and the lang2vec utility \cite{littell2017uriel} has been notable in extending multilingual NLP research, particularly for less-resourced languages, by providing methods to represent and compare the structured lingustic features across different languages. Complementing this, \citet{ponti2017universals} pointed out the underexplored typological features in existing approaches and the need for integrating data-driven methods of typological knowledge into language models. Previous studies have focused on extending multilingual NLP using URIEL vectors \cite{lauscher-etal-2020-zero, lin-etal-2019-choosing, tan-etal-2019-multilingual, oncevay-etal-2020-bridging}, but none of their approaches use URIEL vectors for alignment during the finetuning process of language models.

However, PLMs still face significant challenges in generalizing to unseen languages, particularly when adapting to low-resource and unseen languages. These challenges stem from the vast structural and semantic variation across languages \cite{BenderLinguisticIS, Jurafsky2019SpeechAL}, the scarcity of resources \cite{mohammad2019state, lewis-etal-2020-mlqa}, and the limitations inherent in the models themselves \cite{lin2017structured}. This situation highlights the complexity of generalizing these models effectively to a broader and scope of language. This situation highlights the complexity of generalizing these models effectively to a broader scope of languages, making it difficult for them to perform well on languages they haven't been specifically trained on.

\section{Unseen Languages Adaptation with \methodname{}}

In this section, we provide an overview of how \methodname{} captures linguistic constraints and explain the intuition behind it. We also discuss in detail how we align model representations with the linguistic vectors.

\subsection{Does Multilingual LMs capture Linguistic Constraints?}

We define the linguistic knowledge as a vector gathered from URIEL vector \citep{littell2017uriel}. We chose three distinct linguistic knowledge from the database, namely \texttt{'syntax\_knn'}, \texttt{'syntax\_average'}\footnote{In this work, we chose the \texttt{'knn'} and \texttt{'average} syntax features. These include consensus values (like averages) and predicted values (such as kNN regressions based on phylogenetic or geographical neighbors)}, and \texttt{'geo'} features. The choice of \texttt{'syntax\_knn'} and \texttt{'syntax\_average'} is motivated by the typological nature of syntax. Syntax in languages varies widely; hence, by using aggregate measures like averages and k-nearest neighbors (kNN), we can capture a more general representation of syntactic features across languages. Note that in our experiments, we excluded phonological features and language family attributes from our analysis as they are less relevant to textual data and have limited granularity for understanding linguistic variations.

\paragraph{Syntax Features} These feature vectors denote a typological feature that is adapted from several sources including World Atlas of Language Structures (WALS) \cite{wals}, Syntactic Structures of World Languages \cite{Collins2009SyntacticSO}, and short prose descriptions on typological features in Ethnologue \cite{lewis2009ethnologue}.Syntax vectors capture information about the syntactic properties of languages, derived from large-scale typological databases that document the structural and semantic variation across different languages. These syntax features in URIEL are utilized to represent languages in vector form that allows the analysis and comparison of languages based on their syntactic properties.

\begin{figure}[t!]
    \centerline{\includegraphics[width=\columnwidth]{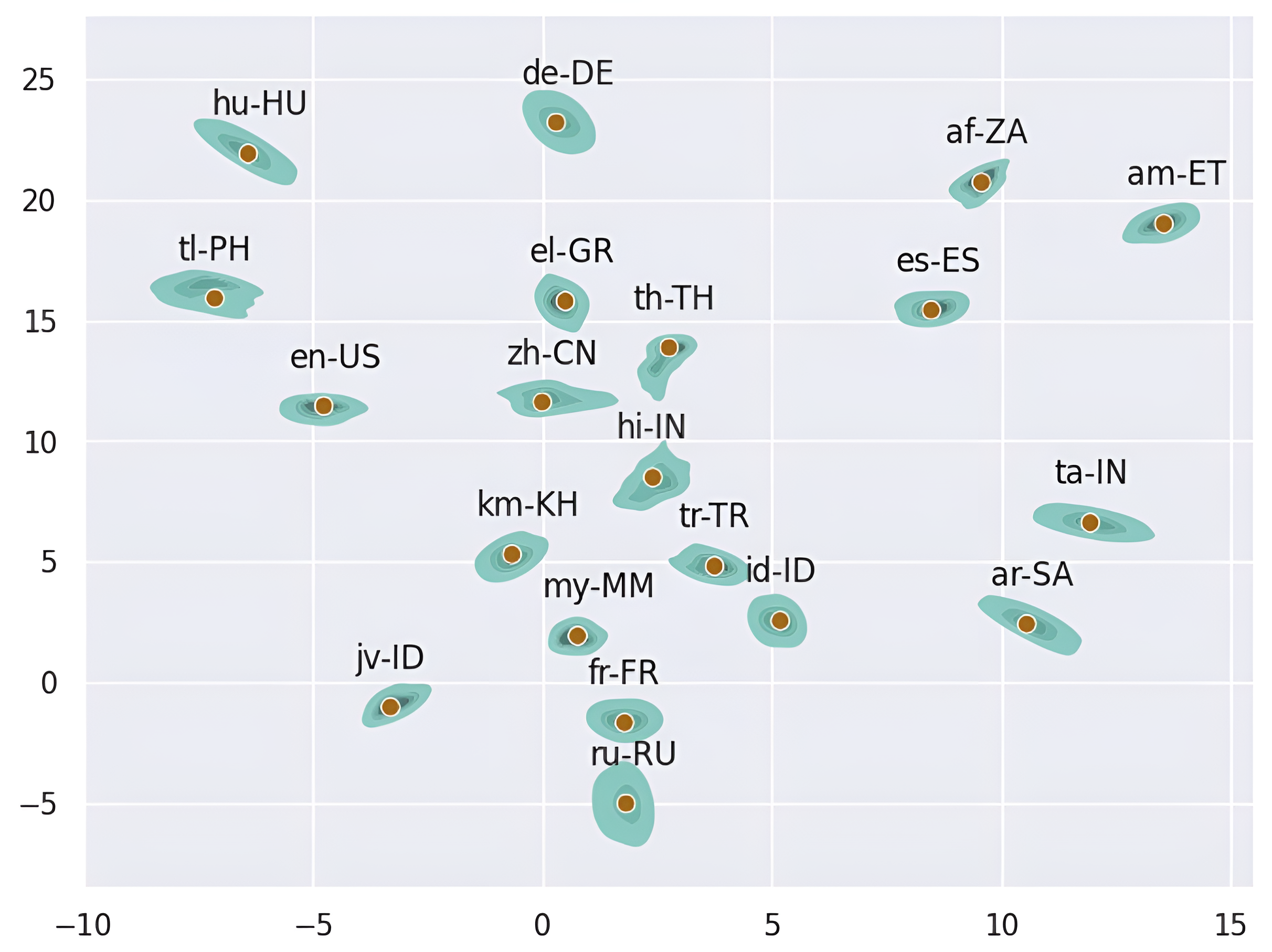}}
    \caption{Alignment between mBERT Representation with URIEL Language Representation. The \textcolor{teal}{green-shaded areas} indicate the sentence representations of mBERT while the \textcolor{brown}{brown dots} represent the URIEL representations of the corresponding language.}
    \label{fig:alignment}
\end{figure}

\paragraph{Geographical Features} On the other hand, geographical features represent languages in terms of their geographical properties.  The inclusion of \texttt{``geo"} features aims to capture geographical attributes of languages. This feature expresses geographical location with a fixed number of dimensions that each represents the ``great circle" distance—from the language in question to a fixed point on the Earth’s surface. By incorporating geographical information into language vectors, URIEL and lang2vec provide a more comprehensive view of languages, considering not only their structural and semantic properties but also their geographical context.

\subsection{Proof of Concept}
\paragraph{Linguistic Separability in LMs} We investigate whether PLMs like mBERT~\cite{devlin2018bert} can capture linguistic constraints by aligning mBERT language embeddings with URIEL vectors to assess how they represent seen and unseen languages. This includes examining how well mBERT’s embeddings correspond to the typological and geographical features detailed in URIEL. In Figure \ref{fig:alignment}, sentence embeddings (\textcolor{teal}{green dots}) from mBERT, derived from the last hidden state of multilingual training data, and URIEL vectors (\textcolor{brown}{brown dots})—structured representations from the URIEL database—are projected into the same space. A matrix \(W\) is used to linearly project sentence embeddings, minimizing the mean squared error with URIEL vectors. This alignment is showcased in Figure \ref{fig:alignment} using UMAP for visualization purpose.

Figure \ref{fig:alignment} presents a  visual analysis facilitated by UMAP~\cite{mcinnes2018umap}, showing the correlation between mBERT language representation and the linguistic vectors from the URIEL database ($R^2$ = 0.816). By leveraging UMAP, the plot highlight the principal variances within the joint feature space of the embeddings and vectors. The spatial representation of languages on this plot mirrors their linguistic and geographical relatedness, as encapsulated by mBERT. This visualization shows the model's ability to mirror linguistic typologies, with languages sharing common roots such as 'de-DE' and 'nl-NL' naturally clustering together. The density and arrangement of these clusters potentially reflect mBERT capacity to capture and represent language family traits. Conversely, the presence of sparser clusters or outliers requires us to carefully check mBERT's coverage and consistency in representing different linguistic features. We also formally defined the language representation alignment in Algorithm~\ref{alg:language-representation-alignment}.

\begin{table}[!ht]
\centering
\small
\begin{tabular}{l l c}
\toprule
\textbf{Model} & \textbf{Feature Type} & \textbf{Acc.(\%)} \\
\midrule
\multirow{7}{*}{mBERT} 
    & geo                                 & 66.14 \\
    & syntax\_avg+geo                 & 65.62 \\
    & syntax\_avg                     & 66.41 \\
    & syntax\_knn+geo                     & 66.08 \\
    & syntax\_knn                         & 66.41 \\
    & \textbf{syntax\_knn+syntax\_avg+geo} & \textbf{66.47} \\
    & syntax\_knn+syntax\_avg         & 66.41 \\
\midrule
\multirow{7}{*}{XLM-R} 
    & geo                                 & 80.16 \\
    & syntax\_avg+geo                 & 80.54 \\
    & syntax\_avg                     & 80.76 \\
    & syntax\_knn+geo                     & 80.48 \\
    & syntax\_knn                         & 80.32 \\
    & \textbf{syntax\_knn+syntax\_avg+geo} & \textbf{80.80} \\
    & syntax\_knn+syntax\_avg         & 80.78 \\
\bottomrule
\end{tabular}
\caption{Linguistic vector ablation experiment (the highest accuracy for each model is highlighted in \textbf{bold}).}
\label{table:vector_ablation}
\end{table}

Table \ref{table:vector_ablation} shows the results of the linguistic vector ablation experiment for mBERT and XLM-R, testing different feature combinations. For mBERT, individual features like syntax\_avg and syntax\_knn perform similarly (~66.41\%), with only minor improvement (66.47\%) when combined with geo. In contrast, XLM-R benefits more from feature combinations, achieving its highest accuracy (80.80\%) when syntax\_knn, syntax\_avg, and geo are combined. Even individually, syntax\_avg (80.76\%) performs well for XLM-R, highlighting the model's stronger ability to leverage syntactic information. These results suggest that combining syntax and geographical features yields optimal performance, especially for XLM-R, and this combination will be used in subsequent experiments.

\subsection{$\vcenter{\hbox{\includegraphics[scale=0.028]{assets/alchemy.png}}}$\methodname}

We introduce \methodname{} as an approach that intuitively aligns model representations with linguistic knowledge, leveraging URIEL vectors. This approach is applied through an auxiliary loss function that is involved in the training process with an added information of linguistic characteristics in the form of URIEL vector.

In \methodname, we enhance the fine-tuning of encoder models such as mBERT for downstream tasks by not only using the regular classification loss but also introducing a novel linguistic regularization term. This is achieved through the implementation of a URIEL loss, designed to align the model’s representations with linguistic knowledge derived from URIEL vectors. Specifically, this process involves applying a linear projection to the model’s pooled output, which aligns it with the URIEL vector space. The URIEL loss is quantified as the mean squared error (MSE) between the projected model outputs and the corresponding URIEL vectors. This dual approach allows for a more linguistically informed model training and increase the model's ability to capture and reflect complex linguistic patterns.
\begin{align}
   \mathcal{L}_{uriel}(Z, U) = \frac{1}{N} \sum_{i=1}^{N} \| Z_i - U_i \|^2,
   \label{eq:uriel_loss} 
\end{align}

\methodname{} is represented by equation \ref{eq:uriel_loss} where $Z$ represents the model-generated representations, $U$ denotes the URIEL vectors, and $N$ is the number of data points. To generate the model representation, we take the output representation from the \( CLS \) token and multiply it with a new, trainable projection layer to transform the vector size so that they are compatible.

Formally, we define the language representation alignment in Algorithm~\ref{alg:language-representation-alignment}, where $F_U$ represents the features extracted from URIEL, $S$ is the set of sentence representations, $H_x$ and $N_x$ are the hidden states and number of attention-masked tokens for a sentence $x$, respectively. The matrix $W$ is used for the linear projection, and $A$ holds the final aligned representations. Algorithm \ref{alg:language-representation-alignment} outlines the process for aligning language representations we use in Figure \ref{fig:alignment}. It leverages the URIEL database for linguistic features, processes sentences through a language model (\(\Theta\)), and aligns these with mBERT representations (\(M\)). The algorithm iteratively updates transformation parameters (\(W\) and \(b\)) through a training loop to minimize the loss between the projected mBERT representations and the target sentence representations in set \(S\), thus achieving aligned language representations (\(A\)).

\begin{algorithm}[!ht]
\caption{Language Representation and Alignment Process}
\label{alg:language-representation-alignment}
\begin{algorithmic} 

\REQUIRE Dataset $D$, URIEL database $U$, Language Model $\Theta$, mBERT representations $M$

\ENSURE Aligned Language Representations $A$

\break

\STATE $F_U \leftarrow$ \textsc{ExtractFeatures}($U$) 
% \Comment{Extraction of URIEL Features}

\STATE $S \leftarrow \{\}$ 
% \Comment{Initialize set for sentence representations}

\FOR{each sentence $x$ in $D$}
    \STATE $H_x \leftarrow$ \textsc{GetLastHiddenStates}($x$, $\Theta$)
    \STATE $N_x \leftarrow$ \textsc{CountAttentionMasked}($x$)
    \STATE $R_x \leftarrow \frac{\textsc{Sum}(H_x)}{N_x}$ 
    % \Comment{Calculate average representation}
    \STATE $S \leftarrow S \cup \{R_x\}$ 
    % \Comment{Add representation to set}
\ENDFOR

\STATE $W, b \leftarrow$ \textsc{InitializeParameters}()
% \Comment{Initialize transformation matrix}
% \Comment{Initialize set for aligned representations}

\FOR{each training epoch}
    % \FOR{each sentence representation $s$ in $S$}
    % \STATE $A_m \leftarrow$ \textsc{AlignWithConstraint}($P_m$, $F_U$)
        % \STATE $A \leftarrow A \cup \{A_m\}$
    % \ENDFOR
    \STATE $P_U \leftarrow (W \times S) + b$
    \STATE $loss \leftarrow$ \textsc{ComputeLoss}($P_U$, $F_U$)
    \STATE $W, b \leftarrow$ \textsc{UpdateParametersWithConstraint}($W, b, loss$)
\ENDFOR

\STATE $A \leftarrow \{\}$ 
\FOR{each sentence representation $s$ in $S$}
    \STATE $A_m \leftarrow (W \times s) + b$
    \STATE $A \leftarrow A \cup \{A_m\}$
\ENDFOR
\end{algorithmic}
\end{algorithm}

Note that there may be discrepancies between the scales of the standard classification loss and the URIEL loss. To address this, we introduce an optional hyperparameter, denoted as $\lambda$, to scale the URIEL loss appropriately. However, finding this scaling factor requires another hyperparameter search. Therefore, we propose a new method for dynamically balancing the classification and URIEL losses using two dynamic scaling approaches.

\paragraph{\textbf{Dynamic Scaling Approaches}} In addition to the fixed scaling factor, we also explore dynamic adjustment of this scaling factor at each training step. This aims to maintain a balance between the classification and URIEL losses, and even considers making the scale trainable. The final loss formula when training with \methodname~is given by:
\begin{align}
   \mathcal{L} = \lambda_{cls} * \mathcal{L}_{cls} + \lambda_{uriel} * \mathcal{L}_{uriel}(Z, U).
\end{align}
We define two methods to implement dynamic scaling:

\begin{enumerate}
    \item \textbf{\textsc{AlchemyScale}}: This method dynamically adjusts the scaling factor \( \lambda \) during training. It is initiated with scaling factors set relative to the mean of initial losses. Furthermore, these factors are updated periodically using an Exponential Moving Average (EMA) method to balance between different loss components.
    
    \item \textbf{\textsc{AlchemyTune}}: Here, \( \lambda \) is conceptualized as a trainable parameter within the model's architecture. Initialized as part of the model's parameters and optimized during the training process. This method applies the scaling factors to loss components, then an additional \textit{mini\_loss} is computed to represent the deviation of the sum of scaling factors.
\end{enumerate}

Both methods aim to enhance model performance by dynamically and intelligently scaling loss components, with the first method relying on predefined, periodically updated scaling mechanisms, and the second integrating the scaling factor into the model's learning parameters for adaptive adjustments.

\section{Experiment Setting}

\paragraph{Datasets} 
In our experiments, we use MASSIVE Dataset~\cite{fitzgerald-etal-2023-massive}, which is a comprehensive collection of multilingual data incorporating intent classification tasks. We split MASSIVE into 25 languages that are ``seen" during finetuning and the rest 27 languages that are ``unseen", which we exclusively used for evaluation. This splitting is based on the language adapters availability as outlined in the prior research of \citet{pfeiffer2020AdapterHub}, which we utilized in the AdapterFusion experiment for our baseline model. For a detailed breakdown of the languages used, including their respective families, genera, and script can be found in Appendix \ref{sec:languages_breakdown}.

Additionally, we incorporate the MasakhaNews Dataset \cite{Adelani2023MasakhaNEWS}, consisting of news article classification across several African languages. This dataset tests our models against diverse journalistic styles and complex syntactic structures. For our experiments, the training languages are \texttt{amh, eng, fra, hau, swa, orm}, and \texttt{som}, while the testing languages include \texttt{ibo, lin, lug, pcm, run, sna, tir, xho}, and \texttt{yor}. Lastly, we also utilize the SemRel2024 Dataset \cite{ousidhoum2024semrel2024} for semantic relatedness task in low-resource languages. We use this dataset to evaluate \methodname{}'s semantic understanding and relationship extraction capabilities. We train using the languages \texttt{amh, arq, ary, eng, esp, hau, kin, mar}, and \texttt{tel}. The test set includes \texttt{afr, amh, arb, arq, ary, eng, esp, hau, hin, ind, kin}, and \texttt{pan}. % \footnote{Full names: Afrikaans (afr), Algerian Arabic (arq), Amharic (amh), English (eng), French (fra), Hausa (hau), Hindi (hin), Igbo (ibo), Indonesian (ind), Kinyarwanda (kin), Lingala (lin), Luganda (lug), Marathi (mar), Moroccan Arabic (ary), Nigerian Pidgin (pcm), Oromo (orm), Punjabi (pan), Rundi (run), Shona (sna), Somali (som), Spanish (esp), Standard Arabic (arb), Swahili (swa), Telugu (tel), Tigrinya (tir), Xhosa (xho), Yoruba (yor).}

\paragraph{Models} 
Our study employs two widely used and resource-efficient multilingual language models: Multilingual BERT Base (mBERT\textsubscript{\textsc{Base}}) and XLM-RoBERTa Base (XLM-R\textsubscript{\textsc{Base}}). In our training process, we use a learning rate of \(5 \times 10^{-5}\), train for 30 epochs, and measure performance based on accuracy for MASSIVE and MasakhaNews, and Pearson correlation for SemRel. Each training takes at most 5 hours using a single A100 GPU.

\begin{table*}[!ht]
\centering
\fontsize{10}{12}\selectfont
\begin{tabular}{@{}>{\centering\arraybackslash}m{1cm} >{\centering\arraybackslash}m{1.5cm}ccccccc@{}}
\toprule
\multirow{2}{*}{\textbf{Model}} & \multirow{2}{1.5cm}{\centering\textbf{Language \\ Category}} & \multicolumn{2}{c}{\textbf{MASSIVE}} & \multicolumn{2}{c}{\textbf{MasakhaNews}} & \multicolumn{2}{c}{\textbf{SemRel}} \\
\cmidrule(lr){3-4} \cmidrule(lr){5-6} \cmidrule(lr){7-8}
 & & \textbf{Base} & \textbf{Ours} & \textbf{Base} & \textbf{Ours} & \textbf{Base} & \textbf{Ours} \\
\midrule
\multirow{3}{*}{\textbf{mBERT}} & \textbf{Low} & 48.43 & 66.93 (\textcolor{darkgreen}{$\uparrow18.50$}) & 46.25 & 72.70 (\textcolor{darkgreen}{$\uparrow26.45$}) & -0.26 & -0.14 (\textcolor{darkgreen}{$\uparrow0.12$}) \\
 & \textbf{Medium} & 55.96 & 64.42 (\textcolor{darkgreen}{$\uparrow8.47$}) & 42.94 & 64.38 (\textcolor{darkgreen}{$\uparrow21.45$}) & 0.16 & 0.21 (\textcolor{darkgreen}{$\uparrow0.05$}) \\
 & \textbf{High} & 79.66 & 66.81 (\textcolor{darkred}{$\downarrow12.85$}) & 77.89 & 73.74 (\textcolor{darkred}{$\downarrow4.16$}) & 0.02 & -0.03 (\textcolor{orange}{$\downarrow0.05$}) \\
\midrule
\multirow{3}{*}{\textbf{XLM-R}} & \textbf{Low} & 78.15 & 80.90 (\textcolor{darkgreen}{$\uparrow2.76$}) & 47.86 & 79.31 (\textcolor{darkgreen}{$\uparrow31.46$}) & 0.08 & 0.40 (\textcolor{darkgreen}{$\uparrow0.32$}) \\
 & \textbf{Medium} & 80.31 & 80.15 (\textcolor{orange}{$\downarrow0.16$}) & 55.87 & 75.79 (\textcolor{darkgreen}{$\uparrow19.92$}) & 0.49 & 0.37 (\textcolor{orange}{$\downarrow0.12$}) \\
 & \textbf{High} & 86.82 & 80.86 (\textcolor{darkred}{$\downarrow5.96$}) & 68.14 & 73.79 (\textcolor{darkgreen}{$\uparrow5.65$}) & 0.26 & 0.41 (\textcolor{darkgreen}{$\uparrow0.15$}) \\
\bottomrule
\end{tabular}
\caption{Performance comparison of mBERT and XLM-R models across different language categories and benchmarks.}
\label{tab:performance_comparison}
\end{table*}

\section{Results and Discussion}

\subsection{\methodname~Performance}
Our results as shown in Table \ref{tab:semrel_masakhanews} reveal that \methodname{} excels across all languages in the MasakhaNews dataset, including those not encountered during the pretraining of mBERT (*) and XLM-R ($\dagger$). \methodname{} further demonstrates notable improvements on the Semantic Relatedness dataset, showing its ability to adapt to languages with distinct typological characteristics from the training corpus. In this experiment, we opted not to compare our method against the baseline used in the Semantic Relatedness paper because LaBSE is not zero-shot; it was pretrained with sentence similarity tasks contrasting our method's conditions. Moreover, we excluded the MAD-X experiment from the MasakhaNews evaluation because MAD-X's parameter-efficient approach differs fundamentally from our full finetuning approach. Collectively, these insights suggest that our \methodname{} can generalize across varied linguistic attributes.

Additionally, we applied the same procedure to MASSIVE dataset and the results are summarized in Table~\ref{tab:massive}. We compared our method with zero-shot generalization, where the model is fully tuned on seen languages and then tested on unseen languages (referred to as Full FT in the Table). Furthermore, we explored AdapterFusion \cite{pfeiffer2021adapterfusion} as another baseline. AdapterFusion has shown better adaptation to unseen languages than naive zero-shot generalization. Unfortunately, many language adapters that we need for AdapterFusion is not available for XLM-R.

From Table~\ref{tab:massive}, it is shown that \methodname~achieves better generalization for unseen languages. We observed a significant improvement for mBERT and a modest average improvement for the stronger XLM-R model. For mBERT, \methodname~can significantly increase performance in truly unseen languages of am-ET, km-KH, mn-MN, in which mBERT has never seen during the pre-training stage nor fine-tuning. These findings show that \methodname~can be useful in truly zero-shot settings. While \methodname~significantly boosts performance in weaker languages such as cy-GB or sw-KE, it can occasionally degrade results in languages with already strong zero-shot performance, particularly evident in XLM-R where it tends to flatten results to the 80-82\% range.

\begin{table}[!ht]
    \centering
    \small
    \renewcommand{\arraystretch}{0.85} % Reduces the height of rows
    \begin{tabular}{llr}
    \toprule
    \textbf{Dataset}      & \textbf{Language} & \textbf{UNK \%} \\
    \midrule
    %\multicolumn{3}{c}{MASSIVE dataset} \\
    %\midrule
      MASSIVE           & am-ET    & 6.79         \\
                 & km-KH    & 3.81         \\
                 & vi-VN    & 0.35         \\
    & {Other languages} & <0.1           \\
    \midrule
    %\multicolumn{3}{c}{SemRel dataset} \\
    %\midrule
     SemRel            & amh      & 3.43         \\
                 & hau      & 0.60         \\
                 & ary      & 0.44         \\
    & {Other languages}  & <0.4        \\
    %\midrule
    %\multicolumn{3}{c}{MasakhaNews dataset} \\
    \midrule
    MasakhaNews             & pcm      & 0.43         \\
                 & eng      & 0.16         \\
    & {Other languages} & <0.1         \\
    \bottomrule
    \end{tabular}
    \caption{UNK percentages in different datasets, illustrating the prevalence of unknown tokens that \methodname~successfully manages.}
    \label{tab:unk_percentages}
\end{table}

Despite the variations in performance, the potential of \methodname{} is particularly clear in scenarios where zero-shot performance is inherently weak. Our hypothesis is that the model indirectly leverages familiar scripts encountered during pretraining and helps its ability to effectively handle UNK tokens. Advances in models using byte-level tokenization units theoretically reduce or eliminate OOV tokens; however, our evaluations across the MASSIVE, MasakhaNews, and SemRel2024 datasets, as shown in Table \ref{tab:unk_percentages}, confirm that UNK tokens have a minimal impact, thus showing the robustness of \methodname{} in such environments. For contexts where UNK token rates are high, the solution might be orthogonal to our approach, requiring further improvement in the base models or tokenizers that could later be integrated with \methodname{}.

\subsection{Effect of Scaling URIEL Loss}

The classification and URIEL losses operate on different scales. Simply adding these losses together would cause the model to prioritize the loss with the larger magnitude. During the early stages of training, we observe that the classification loss is approximately ten times larger than the URIEL loss. In this section, we explore the impact of various scaling factors applied to the URIEL loss.

\begin{figure}[!ht]
    \centering
    \includegraphics[width=\columnwidth]{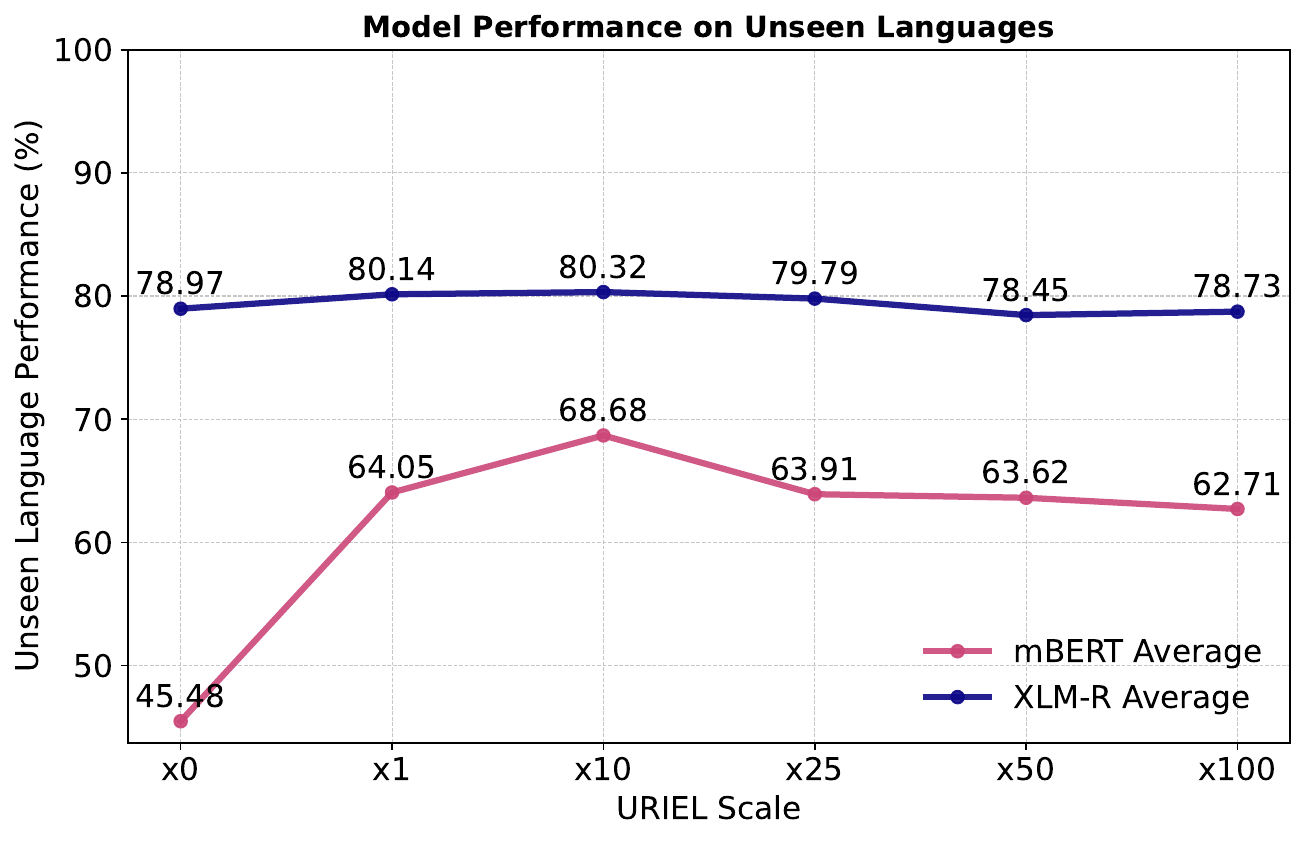}
    \caption{Average performance of unseen languages under various URIEL loss scaling factors.}
    \label{fig:constant-scale}
\end{figure}

\paragraph{Constant Scaling}

We investigate the effect of consistently scaling the URIEL loss by different factors. The results are illustrated in Figure~\ref{fig:constant-scale}. Notably, because we utilize the scale-invariant optimizer AdamW, there is no risk of gradients becoming excessively large due to high loss values. 

As observed in Figure~\ref{fig:constant-scale}, for \textbf{mBERT}, the performance on unseen languages improves significantly when scaling the URIEL loss by 10x, achieving the highest performance of 68.68\%. However, as the scaling factor increases further (e.g., 25x, 50x, and 100x), the performance starts to decline, indicating that overly emphasizing the URIEL loss has diminishing returns or even negative effects. Without any scaling (0x), the model performs poorly, showing the importance of scaling the URIEL loss.

For \textbf{XLM-R}, the trend is more stable. The performance fluctuates slightly across different scaling factors but remains generally consistent, with the highest performance achieved at the 10x scaling factor (80.32\%). Larger scaling factors (e.g., 50x, 100x) do not lead to substantial improvements and may even cause minor drops in performance. This suggests that while scaling helps, XLM-R is less sensitive to the URIEL loss scaling than mBERT. 

Overall, a scaling factor of 10$\times$ appears to give the best balance between classification and URIEL losses.

% Generally, a scaling factor of 10 slightly outperforms other scaling factors, while performance tends to decline with higher scaling factors.

\paragraph{Dynamic and Trainable Scaling}

Introducing a scaling factor adds another tunable hyperparameter, which can complicate the training process. Ideally, we seek a balanced weighting between the classification and URIEL losses. Instead of exhaustively testing various scaling factors, an adaptive scaling approach is more cost-effective and advantageous. Here, we explore two strategies: dynamic scaling and trainable scaling factors. The results of these approaches are presented in Table~\ref{tab:scale}.

\begin{table}[h]
\centering
\small
\begin{tabular}{lcc}
\toprule
\textbf{URIEL Scaling} & \textbf{mBERT} & \textbf{XLM-R} \\
\midrule
Constant 10$\times$     & \textbf{64.68} & 80.32        \\
\textsc{AlchemyScale}    & 62.97          & \textbf{80.43} \\
\textsc{AlchemyTune}     & 63.24          & 79.10        \\
\bottomrule
\end{tabular}
\caption{Performance comparison across different URIEL scaling methods.}
\label{tab:scale}
\end{table}

Interestingly, these dynamic scaling methods do not significantly outperform a constant scaling factor. Specifically, a 10$\times$ scaling achieves the best performance for mBERT, while dynamic scaling only marginally outperforms the 10$\times$ scaling for XLM-R. Therefore, in scenarios with limited computational resources, a 10$\times$ scaling factor is recommended. However, with more computational capacity, exploring different scaling factors may yield marginal gains.

\begin{table*}[!ht]
    \centering
    \begin{minipage}{.4\textwidth}
        \centering
        \includegraphics[width=\linewidth]{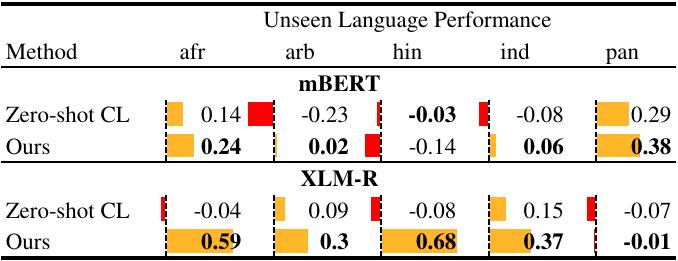} % Example image 1
    \end{minipage}\hfill
    \begin{minipage}{.57\textwidth}
        \centering
        \includegraphics[width=\linewidth, trim={74px 0px 0px 0px}, clip]{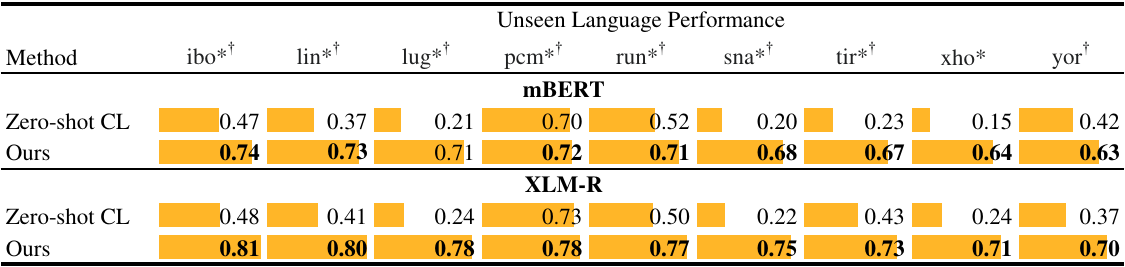} % Example image 2, smaller width
        
    \end{minipage}
    \caption{Performance of \methodname~in SemRel (left) and MasakhaNews (right) dataset for unseen languages. For languages in * and $\dag$, mBERT and XLM-R have never seen the languages during pre-training, respectively.}
    \label{tab:semrel_masakhanews}
\end{table*}

\begin{table*}[!ht]
    \centering
    \centerline{\includegraphics[width=1\linewidth]{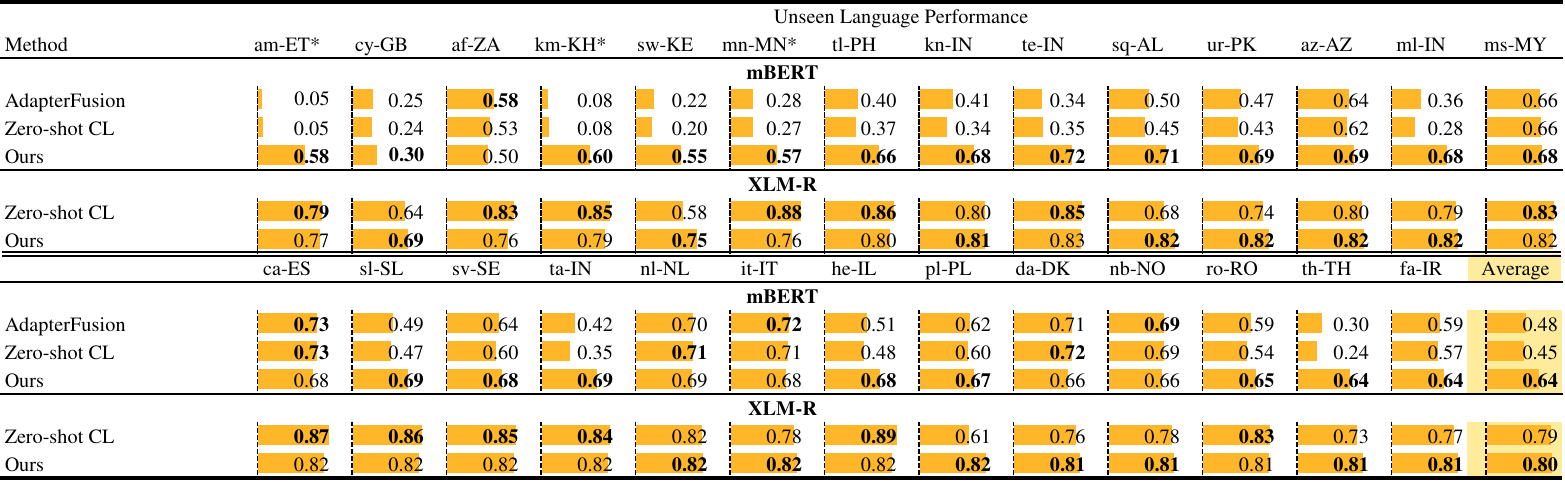}}
    \caption{Performance of \methodname~in MASSIVE dataset for unseen languages. For languages in *, mBERT has never seen the languages during pre-training.}
    \label{tab:massive}
\end{table*}

\subsection{Generalization Across Language Family}
\label{sec:generalization}
We investigate \textsc{\methodname} across language families to further analyze the generalization capabilities of BERT and XLM-R models. We perform our experiment by splitting the languages in MASSIVE according to their language families and train the model on a subset of language families while testing on the rest, unseen language families. We explore on including different subset of language families, as seen in the Appendix (Table~\ref{tab:language_groups}).

As illustrated in Figure 4, \methodname~demonstrates generalization towards these unseen language families. Perhaps unsurprisingly, adding more subset of diverse languages improves generalization performance. Notably, the inclusion of the Afro-Asiatic language group—consisting of languages such as ``am-ET", ``ar-SA", and ``he-IL", each featuring unique scripts—has significantly enhanced performance from the second to the third training group iteration. This improvement underscores \methodname{}'s capability to adapt to scripts not presented during the initial training or fine-tuning phases, illustrating its robustness in generalizing across different scripts.

The performance of both models, combined with \methodname~underscores the advantage of including a broader spectrum of languages within training groups for enhanced model generalization. However, the impact of this diversity is not uniform across all language families: While some consistently benefit from the expansion of training data, others do not, and shows that merely increasing the volume of data from the same family may not necessarily improve performance. This inconsistency highlights the potential limitations within the models' capacity to learn and generalize the linguistic features specific to certain language families. Consequently, our observation shows that the degree of generalization varies among different language families. This suggests that while some may significantly profit from these models' capabilities, others may require more tailored strategies to gain similar performance improvement.

\subsection{Seen Language Performance}
 
While \methodname{} consistently improves performance across unseen languages, we note some inconsistencies concerning the performance of seen languages. In MASSIVE, we observe a noticeable performance drop in seen languages, while in contrast, we still see a massive gain in MasakhaNews and the performance of SemRel seems to be unaffected. The compiled results can be seen in Table~\ref{tab:performance_comparison}.

\begin{figure}[!ht]
    \centerline{\includegraphics[width=0.95\linewidth]{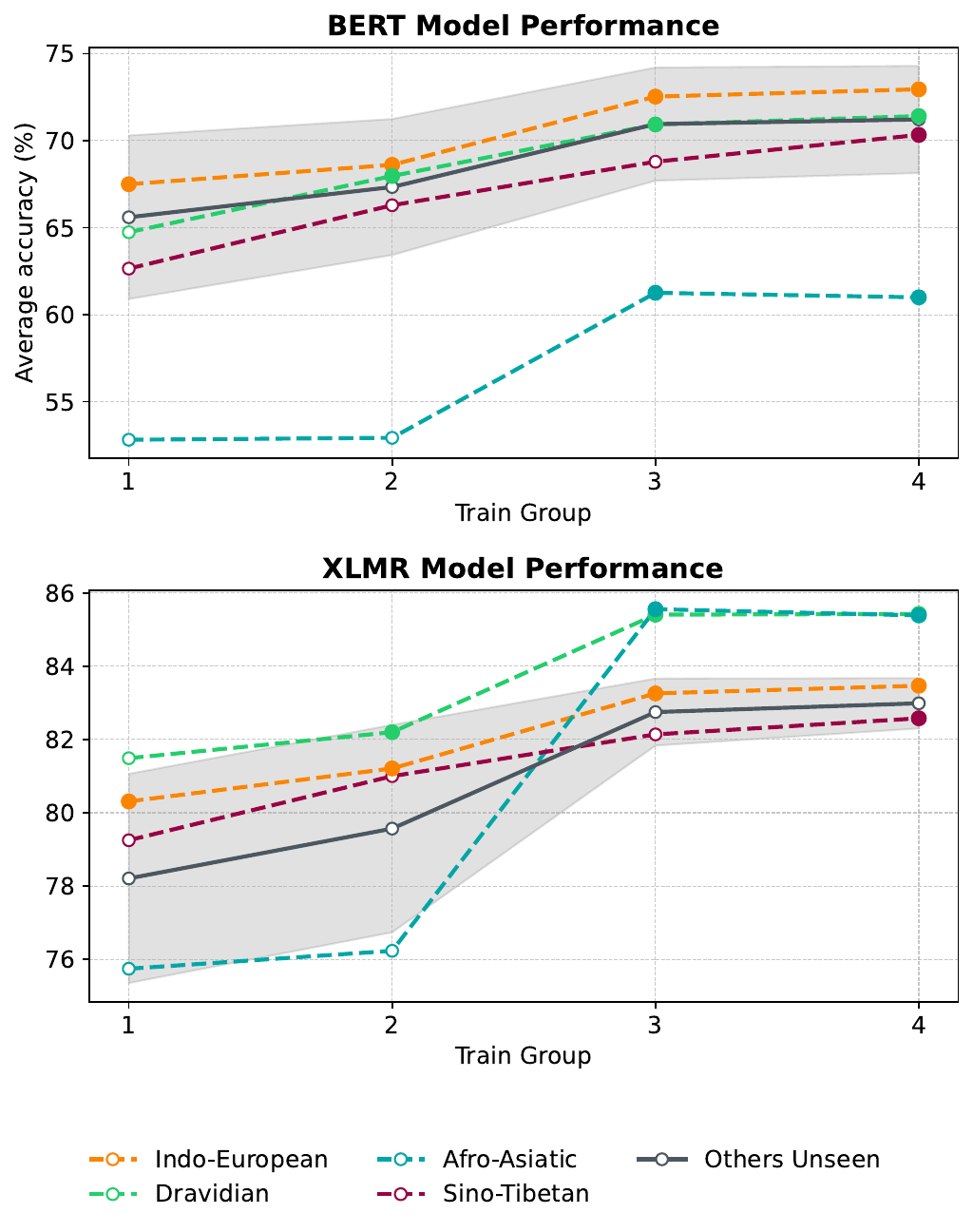}}
    \caption{Model performance across language families.  Dotted lines indicates language families used in training in some of the training stages (solid dots for active use--refer to Table~\ref{tab:language_groups}), and solid grey lines for families unseen in all training stages, with variance shown in shading. 
    }
    \label{fig:lingualchemy-languagefamily}
\end{figure}

In \ref{fig:lingualchemy-languagefamily}, we compare the performance of the BERT and XLMR models across different language families. The dotted lines represent the language families that were included in training (i.e., seen languages), and the solid gray lines represent language families that were unseen in training. Notably, variance for unseen languages is indicated by the shaded areas. This figure shows that while both models generally improve across training groups, the performance of certain language families, particularly seen languages, varies. For instance, in the case of the BERT model, languages from the \textcolor{cyan}{Sino-Tibetan family} demonstrate relatively poor performance across training stages compared to other families, even though they are part of the seen group. Meanwhile, XLMR shows a more consistent performance boost across language families, including unseen ones, though \textcolor{orange}{Indo-European languages} perform better overall.

As MasakhaNews focuses on extremely low-resource languages, we hypothesize that despite being exposed during pretraining, the models' performance remains low even with standard fine-tuning methods. Hence, \methodname can significantly improve performance, even by ~18\% in the low-resource languages. For high-resource languages, traditional fine-tuning is a better choice. We are investigating why \methodname does not help with some languages and how to enhance the performance of some seen languages as part of our future work. Nevertheless, our method still proves beneficial in under-resourced settings where multilingual models typically perform poorly.

\section{Conclusion}
\label{sec:conclusion}
We introduced \methodname, a novel approach that demonstrates strong performance across 30+ unseen languages on intent classification and semantic relatedness tasks. Our method hinges on the integration of linguistic knowledge through the URIEL vectors, enhancing the language model's ability to generalize across a diverse set of languages. We also proposed \textsc{AlchemyScale} and \textsc{AlchemyTune}, which employs a hyperparameter search for the URIEL scaling factor. This is achieved by two key strategies: (1) weight-averaging classification and URIEL loss, and (2) learning to balance the scale between classification and URIEL loss. \methodname{} achieves a massive performance improvement on low-resource languages in multiple downstream tasks including intent classification (\textcolor{darkgreen}{$\uparrow18.50$}), news classification (\textcolor{darkgreen}{$\uparrow31.46$}), and semantic relatedness (\textcolor{darkgreen}{$\uparrow0.32$}).

\section*{Limitations}
\label{sec:limitations}
\methodname{} enhances performance across many unseen languages in intent classification, yet it faces limitations. Performance on seen languages is less than ideal, indicating room for improvement through methods like weight freezing. Also, better generalization appears to reduce accuracy in seen languages, pointing to a need for balanced approaches. Currently, the research is limited to intent classification, and expanding to other NLP tasks could reveal more about its versatility. Moreover, the choice of URIEL features—syntax, geography, language family—is theoretically sound, as discussed in Section 3, but empirical tests with different features might refine the model further. Overcoming these limitations could greatly improve the generalizability and effectiveness of multilingual NLP models.

\bibliography{anthology,custom}

\begin{thebibliography}{45}
\expandafter\ifx\csname natexlab\endcsname\relax\def\natexlab#1{#1}\fi

\bibitem[{Adelani et~al.(2023)Adelani, Masiak, Azime, Alabi, Tonja, Mwase, Ogundepo, Dossou, Oladipo, Nixdorf, Emezue, al~azzawi, Sibanda, David, Ndolela, Mukiibi, Ajayi, Ngoli, Odhiambo, Owodunni, Obiefuna, Muhammad, Abdullahi, Yigezu, Gwadabe, Abdulmumin, Bame, Awoyomi, Shode, Adelani, Kailani, Omotayo, Adeeko, Abeeb, Aremu, Samuel, Siro, Kimotho, Ogbu, Mbonu, Chukwuneke, Fanijo, Ojo, Awosan, Guge, Sari, Nyatsine, Sidume, Yousuf, Oduwole, Kimanuka, Tshinu, Diko, Nxakama, Johar, Gebre, Mohamed, Mohamed, Hassan, Mehamed, Ngabire, , and Stenetorp}]{Adelani2023MasakhaNEWS}
David~Ifeoluwa Adelani, Marek Masiak, Israel~Abebe Azime, Jesujoba~Oluwadara Alabi, Atnafu~Lambebo Tonja, Christine Mwase, Odunayo Ogundepo, Bonaventure F.~P. Dossou, Akintunde Oladipo, Doreen Nixdorf, Chris~Chinenye Emezue, Sana~Sabah al~azzawi, Blessing~K. Sibanda, Davis David, Lolwethu Ndolela, Jonathan Mukiibi, Tunde~Oluwaseyi Ajayi, Tatiana~Moteu Ngoli, Brian Odhiambo, Abraham~Toluwase Owodunni, Nnaemeka~C. Obiefuna, Shamsuddeen~Hassan Muhammad, Saheed~Salahudeen Abdullahi, Mesay~Gemeda Yigezu, Tajuddeen Gwadabe, Idris Abdulmumin, Mahlet~Taye Bame, Oluwabusayo~Olufunke Awoyomi, Iyanuoluwa Shode, Tolulope~Anu Adelani, Habiba~Abdulganiy Kailani, Abdul-Hakeem Omotayo, Adetola Adeeko, Afolabi Abeeb, Anuoluwapo Aremu, Olanrewaju Samuel, Clemencia Siro, Wangari Kimotho, Onyekachi~Raphael Ogbu, Chinedu~E. Mbonu, Chiamaka~I. Chukwuneke, Samuel Fanijo, Jessica Ojo, Oyinkansola~F. Awosan, Tadesse~Kebede Guge, Sakayo~Toadoum Sari, Pamela Nyatsine, Freedmore Sidume, Oreen Yousuf, Mardiyyah Oduwole, Ussen Kimanuka,
  Kanda~Patrick Tshinu, Thina Diko, Siyanda Nxakama, Abdulmejid~Tuni Johar, Sinodos Gebre, Muhidin Mohamed, Shafie~Abdi Mohamed, Fuad~Mire Hassan, Moges~Ahmed Mehamed, Evrard Ngabire, , and Pontus Stenetorp. 2023.
\newblock Masakhanews: News topic classification for african languages.
\newblock \emph{ArXiv}.

\bibitem[{Adilazuarda et~al.(2023)Adilazuarda, Cahyawijaya, and Purwarianti}]{adilazuarda2023obscure}
Muhammad~Farid Adilazuarda, Samuel Cahyawijaya, and Ayu Purwarianti. 2023.
\newblock \href {http://arxiv.org/abs/2311.12375} {The obscure limitation of modular multilingual language models}.

\bibitem[{Alabi et~al.(2022)Alabi, Adelani, Mosbach, and Klakow}]{alabi2022adapting}
Jesujoba~O. Alabi, David~Ifeoluwa Adelani, Marius Mosbach, and Dietrich Klakow. 2022.
\newblock \href {http://arxiv.org/abs/2204.06487} {Adapting pre-trained language models to african languages via multilingual adaptive fine-tuning}.

\bibitem[{Ansell et~al.(2021)Ansell, Ponti, Pfeiffer, Ruder, Glava{\v{s}}, Vuli{\'c}, and Korhonen}]{ansell-etal-2021-mad-g}
Alan Ansell, Edoardo~Maria Ponti, Jonas Pfeiffer, Sebastian Ruder, Goran Glava{\v{s}}, Ivan Vuli{\'c}, and Anna Korhonen. 2021.
\newblock \href {https://doi.org/10.18653/v1/2021.findings-emnlp.410} {{MAD}-{G}: {M}ultilingual adapter generation for efficient cross-lingual transfer}.
\newblock In \emph{Findings of the Association for Computational Linguistics: EMNLP 2021}, pages 4762--4781, Punta Cana, Dominican Republic. Association for Computational Linguistics.

\bibitem[{Bender(2011)}]{BenderLinguisticIS}
Emily~M. Bender. 2011.
\newblock \href {https://doi.org/10.33011/lilt.v6i.1239} {Linguistic issues in language technology on achieving and evaluating language-independence in nlp}.
\newblock \emph{Linguistic Issues in Language Technology}, 6.

\bibitem[{Cahyawijaya et~al.(2024)Cahyawijaya, Lovenia, Koto, Putri, Dave, Lee, Shadieq, Cenggoro, Akbar, Mahendra et~al.}]{cahyawijaya2024cendol}
Samuel Cahyawijaya, Holy Lovenia, Fajri Koto, Rifki~Afina Putri, Emmanuel Dave, Jhonson Lee, Nuur Shadieq, Wawan Cenggoro, Salsabil~Maulana Akbar, Muhammad~Ihza Mahendra, et~al. 2024.
\newblock Cendol: Open instruction-tuned generative large language models for indonesian languages.
\newblock \emph{arXiv preprint arXiv:2404.06138}.

\bibitem[{Collins(2010)}]{Collins2009SyntacticSO}
Chris Collins. 2010.
\newblock \href {https://ling.yale.edu/syntactic-structures-worlds-language-cross-linguistic-database} {Syntactic structures of the world's languages (sswl)}.
\newblock Colloquium presented at Yale University.

\bibitem[{Conneau et~al.(2020)Conneau, Khandelwal, Goyal, Chaudhary, Wenzek, Guzmán, Grave, Ott, Zettlemoyer, and Stoyanov}]{conneau2020unsupervised}
Alexis Conneau, Kartikay Khandelwal, Naman Goyal, Vishrav Chaudhary, Guillaume Wenzek, Francisco Guzmán, Edouard Grave, Myle Ott, Luke Zettlemoyer, and Veselin Stoyanov. 2020.
\newblock \href {http://arxiv.org/abs/1911.02116} {Unsupervised cross-lingual representation learning at scale}.

\bibitem[{Devlin et~al.(2018)Devlin, Chang, Lee, and Toutanova}]{devlin2018bert}
Jacob Devlin, Ming-Wei Chang, Kenton Lee, and Kristina Toutanova. 2018.
\newblock Bert: Pre-training of deep bidirectional transformers for language understanding.
\newblock \emph{arXiv preprint arXiv:1810.04805}.

\bibitem[{Devlin et~al.(2019)Devlin, Chang, Lee, and Toutanova}]{devlin-etal-2019-bert}
Jacob Devlin, Ming-Wei Chang, Kenton Lee, and Kristina Toutanova. 2019.
\newblock \href {https://doi.org/10.18653/v1/N19-1423} {{BERT}: Pre-training of deep bidirectional transformers for language understanding}.
\newblock In \emph{Proceedings of the 2019 Conference of the North {A}merican Chapter of the Association for Computational Linguistics: Human Language Technologies, Volume 1 (Long and Short Papers)}, pages 4171--4186, Minneapolis, Minnesota. Association for Computational Linguistics.

\bibitem[{Dryer and Haspelmath(2013)}]{wals}
Matthew~S. Dryer and Martin Haspelmath, editors. 2013.
\newblock \href {https://doi.org/10.5281/zenodo.7385533} {\emph{WALS Online (v2020.3)}}.
\newblock Zenodo.

\bibitem[{Ebrahimi et~al.(2022)Ebrahimi, Mager, Oncevay, Chaudhary, Chiruzzo, Fan, Ortega, Ramos, Rios, Meza-Ruiz, Giménez-Lugo, Mager, Neubig, Palmer, Coto-Solano, Vu, and Kann}]{ebrahimi2022americasnli}
Abteen Ebrahimi, Manuel Mager, Arturo Oncevay, Vishrav Chaudhary, Luis Chiruzzo, Angela Fan, John Ortega, Ricardo Ramos, Annette Rios, Ivan Meza-Ruiz, Gustavo~A. Giménez-Lugo, Elisabeth Mager, Graham Neubig, Alexis Palmer, Rolando Coto-Solano, Ngoc~Thang Vu, and Katharina Kann. 2022.
\newblock \href {http://arxiv.org/abs/2104.08726} {Americasnli: Evaluating zero-shot natural language understanding of pretrained multilingual models in truly low-resource languages}.

\bibitem[{FitzGerald et~al.(2023)FitzGerald, Hench, Peris, Mackie, Rottmann, Sanchez, Nash, Urbach, Kakarala, Singh, Ranganath, Crist, Britan, Leeuwis, Tur, and Natarajan}]{fitzgerald-etal-2023-massive}
Jack FitzGerald, Christopher Hench, Charith Peris, Scott Mackie, Kay Rottmann, Ana Sanchez, Aaron Nash, Liam Urbach, Vishesh Kakarala, Richa Singh, Swetha Ranganath, Laurie Crist, Misha Britan, Wouter Leeuwis, Gokhan Tur, and Prem Natarajan. 2023.
\newblock \href {https://doi.org/10.18653/v1/2023.acl-long.235} {{MASSIVE}: A 1{M}-example multilingual natural language understanding dataset with 51 typologically-diverse languages}.
\newblock In \emph{Proceedings of the 61st Annual Meeting of the Association for Computational Linguistics (Volume 1: Long Papers)}, pages 4277--4302, Toronto, Canada. Association for Computational Linguistics.

\bibitem[{Ganesh et~al.(2021)Ganesh, Chen, Lou, Khan, Yang, Sajjad, Nakov, Chen, and Winslett}]{ganesh2021transformers}
Prakhar Ganesh, Yao Chen, Xin Lou, Mohammad~Ali Khan, Yin Yang, Hassan Sajjad, Preslav Nakov, Deming Chen, and Marianne Winslett. 2021.
\newblock \href {https://doi.org/10.1162/tacl_a_00413} {{Compressing Large-Scale Transformer-Based Models: A Case Study on BERT}}.
\newblock \emph{Transactions of the Association for Computational Linguistics}, 9:1061--1080.

\bibitem[{Goyal et~al.(2021)Goyal, Du, Ott, Anantharaman, and Conneau}]{goyal2021largerscale}
Naman Goyal, Jingfei Du, Myle Ott, Giri Anantharaman, and Alexis Conneau. 2021.
\newblock \href {http://arxiv.org/abs/2105.00572} {Larger-scale transformers for multilingual masked language modeling}.

\bibitem[{Jurafsky and Martin(2019)}]{Jurafsky2019SpeechAL}
Daniel Jurafsky and James~H. Martin. 2019.
\newblock \href {https://web.stanford.edu/~jurafsky/slp3/old_oct19/17.pdf} {Speech and language processing}.

\bibitem[{Kakwani et~al.(2020)Kakwani, Kunchukuttan, Golla, Gokul, Bhattacharyya, Khapra, and Kumar}]{kakwani2020indicnlpsuite}
Divyanshu Kakwani, Anoop Kunchukuttan, Satish Golla, NC~Gokul, Avik Bhattacharyya, Mitesh~M Khapra, and Pratyush Kumar. 2020.
\newblock Indicnlpsuite: Monolingual corpora, evaluation benchmarks and pre-trained multilingual language models for indian languages.
\newblock In \emph{Findings of the Association for Computational Linguistics: EMNLP 2020}, pages 4948--4961.

\bibitem[{Lauscher et~al.(2020)Lauscher, Ravishankar, Vuli{\'c}, and Glava{\v{s}}}]{lauscher-etal-2020-zero}
Anne Lauscher, Vinit Ravishankar, Ivan Vuli{\'c}, and Goran Glava{\v{s}}. 2020.
\newblock \href {https://doi.org/10.18653/v1/2020.emnlp-main.363} {From zero to hero: {O}n the limitations of zero-shot language transfer with multilingual {T}ransformers}.
\newblock In \emph{Proceedings of the 2020 Conference on Empirical Methods in Natural Language Processing (EMNLP)}, pages 4483--4499, Online. Association for Computational Linguistics.

\bibitem[{Lewis(2009)}]{lewis2009ethnologue}
Melvyn Lewis. 2009.
\newblock \emph{Ethnologue: Languages of the World}, volume~9.
\newblock SIL International.

\bibitem[{Lewis et~al.(2019)Lewis, Liu, Goyal, Ghazvininejad, Mohamed, Levy, Stoyanov, and Zettlemoyer}]{lewis2019bart}
Mike Lewis, Yinhan Liu, Naman Goyal, Marjan Ghazvininejad, Abdelrahman Mohamed, Omer Levy, Ves Stoyanov, and Luke Zettlemoyer. 2019.
\newblock \href {http://arxiv.org/abs/1910.13461} {Bart: Denoising sequence-to-sequence pre-training for natural language generation, translation, and comprehension}.

\bibitem[{Lewis et~al.(2020)Lewis, Oguz, Rinott, Riedel, and Schwenk}]{lewis-etal-2020-mlqa}
Patrick Lewis, Barlas Oguz, Ruty Rinott, Sebastian Riedel, and Holger Schwenk. 2020.
\newblock \href {https://doi.org/10.18653/v1/2020.acl-main.653} {{MLQA}: Evaluating cross-lingual extractive question answering}.
\newblock In \emph{Proceedings of the 58th Annual Meeting of the Association for Computational Linguistics}, pages 7315--7330, Online. Association for Computational Linguistics.

\bibitem[{Li et~al.(2021)Li, Tang, Zhao, and Wen}]{li2021pretrained}
Junyi Li, Tianyi Tang, Wayne~Xin Zhao, and Ji-Rong Wen. 2021.
\newblock \href {http://arxiv.org/abs/2105.10311} {Pretrained language models for text generation: A survey}.

\bibitem[{Lin et~al.(2019)Lin, Chen, Lee, Li, Zhang, Xia, Rijhwani, He, Zhang, Ma, Anastasopoulos, Littell, and Neubig}]{lin-etal-2019-choosing}
Yu-Hsiang Lin, Chian-Yu Chen, Jean Lee, Zirui Li, Yuyan Zhang, Mengzhou Xia, Shruti Rijhwani, Junxian He, Zhisong Zhang, Xuezhe Ma, Antonios Anastasopoulos, Patrick Littell, and Graham Neubig. 2019.
\newblock \href {https://doi.org/10.18653/v1/P19-1301} {Choosing transfer languages for cross-lingual learning}.
\newblock In \emph{Proceedings of the 57th Annual Meeting of the Association for Computational Linguistics}, pages 3125--3135, Florence, Italy. Association for Computational Linguistics.

\bibitem[{Lin et~al.(2017)Lin, Feng, dos Santos, Yu, Xiang, Zhou, and Bengio}]{lin2017structured}
Zhouhan Lin, Minwei Feng, Cicero~Nogueira dos Santos, Mo~Yu, Bing Xiang, Bowen Zhou, and Yoshua Bengio. 2017.
\newblock \href {http://arxiv.org/abs/1703.03130} {A structured self-attentive sentence embedding}.

\bibitem[{Littell et~al.(2017)Littell, Mortensen, Lin, Kairis, Turner, and Levin}]{littell2017uriel}
Patrick Littell, David~R. Mortensen, Ke~Lin, Katherine Kairis, Carlisle Turner, and Lori Levin. 2017.
\newblock \href {http://aclweb.org/anthology/E17-2002} {Uriel and lang2vec: Representing languages as typological, geographical, and phylogenetic vectors}.
\newblock In \emph{Proceedings of the 15th Conference of the European Chapter of the Association for Computational Linguistics: Volume 2, Short Papers}, pages 8--14. Association for Computational Linguistics.

\bibitem[{Liu et~al.(2019)Liu, Ott, Goyal, Du, Joshi, Chen, Levy, Lewis, Zettlemoyer, and Stoyanov}]{liu2019roberta}
Yinhan Liu, Myle Ott, Naman Goyal, Jingfei Du, Mandar Joshi, Danqi Chen, Omer Levy, Mike Lewis, Luke Zettlemoyer, and Veselin Stoyanov. 2019.
\newblock \href {http://arxiv.org/abs/1907.11692} {Roberta: A robustly optimized bert pretraining approach}.

\bibitem[{Martin et~al.(2020)Martin, Muller, Suarez, Dupont, Romary, De~La~Clergerie, Seddah, and Sagot}]{martin2020camembert}
Louis Martin, Benjamin Muller, Pedro~Ortiz Suarez, Yoann Dupont, Laurent Romary, {\'E}ric~Villemonte De~La~Clergerie, Djam{\'e} Seddah, and Beno{\^\i}t Sagot. 2020.
\newblock Camembert: a tasty french language model.
\newblock In \emph{Proceedings of the 58th Annual Meeting of the Association for Computational Linguistics}, pages 7203--7219.

\bibitem[{McInnes et~al.(2018)McInnes, Healy, Saul, and Großberger}]{mcinnes2018umap}
Leland McInnes, John Healy, Nathaniel Saul, and Lukas Großberger. 2018.
\newblock \href {https://doi.org/10.21105/joss.00861} {Umap: Uniform manifold approximation and projection}.
\newblock \emph{Journal of Open Source Software}, 3(29):861.

\bibitem[{Mohammad(2019)}]{mohammad2019state}
Saif~M. Mohammad. 2019.
\newblock \href {http://arxiv.org/abs/1911.03562} {The state of nlp literature: A diachronic analysis of the acl anthology}.

\bibitem[{Oncevay et~al.(2020)Oncevay, Haddow, and Birch}]{oncevay-etal-2020-bridging}
Arturo Oncevay, Barry Haddow, and Alexandra Birch. 2020.
\newblock \href {https://doi.org/10.18653/v1/2020.emnlp-main.187} {Bridging linguistic typology and multilingual machine translation with multi-view language representations}.
\newblock In \emph{Proceedings of the 2020 Conference on Empirical Methods in Natural Language Processing (EMNLP)}, pages 2391--2406, Online. Association for Computational Linguistics.

\bibitem[{Ousidhoum et~al.(2024)Ousidhoum, Muhammad, Abdalla, Abdulmumin, Ahmad, Ahuja, Aji, Araujo, Ayele, Baswani, Beloucif, Biemann, Bourhim, Kock, Dekebo, Hourrane, Kanumolu, Madasu, Rutunda, Shrivastava, Solorio, Surange, Tilaye, Vishnubhotla, Winata, Yimam, and Mohammad}]{ousidhoum2024semrel2024}
Nedjma Ousidhoum, Shamsuddeen~Hassan Muhammad, Mohamed Abdalla, Idris Abdulmumin, Ibrahim~Said Ahmad, Sanchit Ahuja, Alham~Fikri Aji, Vladimir Araujo, Abinew~Ali Ayele, Pavan Baswani, Meriem Beloucif, Chris Biemann, Sofia Bourhim, Christine~De Kock, Genet~Shanko Dekebo, Oumaima Hourrane, Gopichand Kanumolu, Lokesh Madasu, Samuel Rutunda, Manish Shrivastava, Thamar Solorio, Nirmal Surange, Hailegnaw~Getaneh Tilaye, Krishnapriya Vishnubhotla, Genta Winata, Seid~Muhie Yimam, and Saif~M. Mohammad. 2024.
\newblock \href {http://arxiv.org/abs/2402.08638} {Semrel2024: A collection of semantic textual relatedness datasets for 14 languages}.

\bibitem[{Pfeiffer et~al.(2021{\natexlab{a}})Pfeiffer, Kamath, Rücklé, Cho, and Gurevych}]{pfeiffer2021adapterfusion}
Jonas Pfeiffer, Aishwarya Kamath, Andreas Rücklé, Kyunghyun Cho, and Iryna Gurevych. 2021{\natexlab{a}}.
\newblock \href {http://arxiv.org/abs/2005.00247} {Adapterfusion: Non-destructive task composition for transfer learning}.

\bibitem[{Pfeiffer et~al.(2020{\natexlab{a}})Pfeiffer, R\"uckl\'{e}, Poth, Kamath, Vuli\'{c}, Ruder, Cho, and Gurevych}]{pfeiffer2020AdapterHub}
Jonas Pfeiffer, Andreas R\"uckl\'{e}, Clifton Poth, Aishwarya Kamath, Ivan Vuli\'{c}, Sebastian Ruder, Kyunghyun Cho, and Iryna Gurevych. 2020{\natexlab{a}}.
\newblock \href {https://www.aclweb.org/anthology/2020.emnlp-demos.7} {Adapterhub: A framework for adapting transformers}.
\newblock In \emph{Proceedings of the 2020 Conference on Empirical Methods in Natural Language Processing (EMNLP 2020): Systems Demonstrations}, pages 46--54, Online. Association for Computational Linguistics.

\bibitem[{Pfeiffer et~al.(2020{\natexlab{b}})Pfeiffer, Vuli{\'c}, Gurevych, and Ruder}]{pfeiffer-etal-2020-mad}
Jonas Pfeiffer, Ivan Vuli{\'c}, Iryna Gurevych, and Sebastian Ruder. 2020{\natexlab{b}}.
\newblock \href {https://doi.org/10.18653/v1/2020.emnlp-main.617} {{MAD-X}: {A}n {A}dapter-{B}ased {F}ramework for {M}ulti-{T}ask {C}ross-{L}ingual {T}ransfer}.
\newblock In \emph{Proceedings of the 2020 Conference on Empirical Methods in Natural Language Processing (EMNLP)}, pages 7654--7673, Online. Association for Computational Linguistics.

\bibitem[{Pfeiffer et~al.(2021{\natexlab{b}})Pfeiffer, Vuli{\'c}, Gurevych, and Ruder}]{pfeiffer-etal-2021-unks}
Jonas Pfeiffer, Ivan Vuli{\'c}, Iryna Gurevych, and Sebastian Ruder. 2021{\natexlab{b}}.
\newblock \href {https://doi.org/10.18653/v1/2021.emnlp-main.800} {{UNK}s everywhere: {A}dapting multilingual language models to new scripts}.
\newblock In \emph{Proceedings of the 2021 Conference on Empirical Methods in Natural Language Processing}, pages 10186--10203, Online and Punta Cana, Dominican Republic. Association for Computational Linguistics.

\bibitem[{Ponti et~al.(2019)Ponti, O’Horan, Berzak, Vulić, Reichart, Poibeau, Shutova, and Korhonen}]{ponti2017universals}
Edoardo~Maria Ponti, Helen O’Horan, Yevgeni Berzak, Ivan Vulić, Roi Reichart, Thierry Poibeau, Ekaterina Shutova, and Anna Korhonen. 2019.
\newblock \href {https://doi.org/10.1162/coli_a_00357} {{Modeling Language Variation and Universals: A Survey on Typological Linguistics for Natural Language Processing}}.
\newblock \emph{Computational Linguistics}, 45(3):559--601.

\bibitem[{Raffel et~al.(2023)Raffel, Shazeer, Roberts, Lee, Narang, Matena, Zhou, Li, and Liu}]{raffel2023exploring}
Colin Raffel, Noam Shazeer, Adam Roberts, Katherine Lee, Sharan Narang, Michael Matena, Yanqi Zhou, Wei Li, and Peter~J. Liu. 2023.
\newblock \href {http://arxiv.org/abs/1910.10683} {Exploring the limits of transfer learning with a unified text-to-text transformer}.

\bibitem[{Rathore et~al.(2023)Rathore, Dhingra, Singla, and {Mausam}}]{rathore-etal-2023-zgul}
Vipul Rathore, Rajdeep Dhingra, Parag Singla, and {Mausam}. 2023.
\newblock \href {https://doi.org/10.18653/v1/2023.emnlp-main.431} {{ZGUL}: Zero-shot generalization to unseen languages using multi-source ensembling of language adapters}.
\newblock In \emph{Proceedings of the 2023 Conference on Empirical Methods in Natural Language Processing}, pages 6969--6987, Singapore. Association for Computational Linguistics.

\bibitem[{Sanh et~al.(2022)Sanh, Webson, Raffel, Bach, Sutawika, Alyafeai, Chaffin, Stiegler, Scao, Raja, Dey, Bari, Xu, Thakker, Sharma, Szczechla, Kim, Chhablani, Nayak, Datta, Chang, Jiang, Wang, Manica, Shen, Yong, Pandey, Bawden, Wang, Neeraj, Rozen, Sharma, Santilli, Fevry, Fries, Teehan, Bers, Biderman, Gao, Wolf, and Rush}]{sanh2022multitask}
Victor Sanh, Albert Webson, Colin Raffel, Stephen~H. Bach, Lintang Sutawika, Zaid Alyafeai, Antoine Chaffin, Arnaud Stiegler, Teven~Le Scao, Arun Raja, Manan Dey, M~Saiful Bari, Canwen Xu, Urmish Thakker, Shanya~Sharma Sharma, Eliza Szczechla, Taewoon Kim, Gunjan Chhablani, Nihal Nayak, Debajyoti Datta, Jonathan Chang, Mike Tian-Jian Jiang, Han Wang, Matteo Manica, Sheng Shen, Zheng~Xin Yong, Harshit Pandey, Rachel Bawden, Thomas Wang, Trishala Neeraj, Jos Rozen, Abheesht Sharma, Andrea Santilli, Thibault Fevry, Jason~Alan Fries, Ryan Teehan, Tali Bers, Stella Biderman, Leo Gao, Thomas Wolf, and Alexander~M. Rush. 2022.
\newblock \href {http://arxiv.org/abs/2110.08207} {Multitask prompted training enables zero-shot task generalization}.

\bibitem[{Tan et~al.(2019)Tan, Chen, He, Xia, Qin, and Liu}]{tan-etal-2019-multilingual}
Xu~Tan, Jiale Chen, Di~He, Yingce Xia, Tao Qin, and Tie-Yan Liu. 2019.
\newblock \href {https://doi.org/10.18653/v1/D19-1089} {Multilingual neural machine translation with language clustering}.
\newblock In \emph{Proceedings of the 2019 Conference on Empirical Methods in Natural Language Processing and the 9th International Joint Conference on Natural Language Processing (EMNLP-IJCNLP)}, pages 963--973, Hong Kong, China. Association for Computational Linguistics.

\bibitem[{{\"U}st{\"u}n et~al.(2020){\"U}st{\"u}n, Bisazza, Bouma, and van Noord}]{ustun-etal-2020-udapter}
Ahmet {\"U}st{\"u}n, Arianna Bisazza, Gosse Bouma, and Gertjan van Noord. 2020.
\newblock \href {https://doi.org/10.18653/v1/2020.emnlp-main.180} {{UD}apter: Language adaptation for truly {U}niversal {D}ependency parsing}.
\newblock In \emph{Proceedings of the 2020 Conference on Empirical Methods in Natural Language Processing (EMNLP)}, pages 2302--2315, Online. Association for Computational Linguistics.

\bibitem[{{\"U}st{\"u}n et~al.(2022){\"U}st{\"u}n, Bisazza, Bouma, and van Noord}]{ustun-etal-2022-udapter}
Ahmet {\"U}st{\"u}n, Arianna Bisazza, Gosse Bouma, and Gertjan van Noord. 2022.
\newblock \href {https://doi.org/10.1162/coli_a_00443} {{UD}apter: Typology-based language adapters for multilingual dependency parsing and sequence labeling}.
\newblock \emph{Computational Linguistics}, 48(3):555--592.

\bibitem[{Wilie et~al.(2020)Wilie, Vincentio, Winata, Cahyawijaya, Li, Lim, Soleman, Mahendra, Fung, Bahar, and Purwarianti}]{wilie-etal-2020-indonlu}
Bryan Wilie, Karissa Vincentio, Genta~Indra Winata, Samuel Cahyawijaya, Xiaohong Li, Zhi~Yuan Lim, Sidik Soleman, Rahmad Mahendra, Pascale Fung, Syafri Bahar, and Ayu Purwarianti. 2020.
\newblock \href {https://aclanthology.org/2020.aacl-main.85} {{I}ndo{NLU}: Benchmark and resources for evaluating {I}ndonesian natural language understanding}.
\newblock In \emph{Proceedings of the 1st Conference of the Asia-Pacific Chapter of the Association for Computational Linguistics and the 10th International Joint Conference on Natural Language Processing}, pages 843--857, Suzhou, China. Association for Computational Linguistics.

\bibitem[{Xu et~al.(2022)Xu, Jonnalagadda, and Durrett}]{xu-etal-2022-massive}
Jiacheng Xu, Siddhartha Jonnalagadda, and Greg Durrett. 2022.
\newblock \href {https://doi.org/10.18653/v1/2022.naacl-main.344} {Massive-scale decoding for text generation using lattices}.
\newblock In \emph{Proceedings of the 2022 Conference of the North American Chapter of the Association for Computational Linguistics: Human Language Technologies}, pages 4659--4676, Seattle, United States. Association for Computational Linguistics.

\bibitem[{Yong et~al.(2023)Yong, Schoelkopf, Muennighoff, Aji, Adelani, Almubarak, Bari, Sutawika, Kasai, Baruwa et~al.}]{yong2023bloom+}
Zheng~Xin Yong, Hailey Schoelkopf, Niklas Muennighoff, Alham~Fikri Aji, David~Ifeoluwa Adelani, Khalid Almubarak, M~Saiful Bari, Lintang Sutawika, Jungo Kasai, Ahmed Baruwa, et~al. 2023.
\newblock Bloom+ 1: Adding language support to bloom for zero-shot prompting.
\newblock In \emph{Proceedings of the 61st Annual Meeting of the Association for Computational Linguistics (Volume 1: Long Papers)}, pages 11682--11703.

\end{thebibliography}
\bibliographystyle{acl_natbib}

\clearpage
\appendix

\section{Languages in Dataset}
\label{sec:languages_breakdown}

The MASSIVE \textit{Dataset}, also known as the \textit{Multilingual Amazon SLU Resource Package} (SLUPR), offers a comprehensive collection of approximately one million annotated utterances for various natural language understanding tasks such as slot-filling, intent detection, and Virtual Assistant performance evaluation. It is an extensive dataset that includes 51 languages, 60 intents, 55 slot types, and spans 18 different domains. The dataset is further enriched with a substantial amount of English seed data, comprising 587k training utterances, 104k development utterances, and 152k test utterances.

\begin{table*}[!h]
\centering
\small
\begin{tabular}{clp{0.5\textwidth}c}
\toprule
\textbf{Train Group} & \textbf{Lang. Family} & \centering\textbf{Languages} & \textbf{Num. Languages} \\
\midrule
1 & Indo-European & af-ZA, bn-BD, ca-ES, cy-GB, da-DK, de-DE, el-GR, en-US, es-ES, fa-IR, fr-FR, hi-IN, hy-AM, is-IS, it-IT, lv-LV, nb-NO, nl-NL, pl-PL, pt-PT, ro-RO, ru-RU, sl-SL, sq-AL, sv-SE, ur-PK & 26 \\
2 & Dravidian & Train Group 1 + kn-IN, ml-IN, ta-IN, te-IN & 30 \\
3 & Afro-Asiatic & Train Group 2 + am-ET, ar-SA, he-IL & 33 \\
4 & Sino-Tibetan & Train Group 3 + my-MM, zh-CN, zh-TW & 36 \\
\midrule
\multicolumn{2}{c}{\multirow{2}{*}{Unseen Languages}} & sw-KE, km-KH, vi-VN, id-ID, jv-ID, ms-MY, tl-PH, ja-JP, ka-GE, ko-KR, mn-MN, th-TH, az-AZ, tr-TR, fi-FI, hu-HU & \multirow{2}{*}{16}\\
\bottomrule
\end{tabular}
\caption{Language family distribution used in the language family generalization experiment (\S\ref{sec:generalization}). The "others unseen" category includes additional language families not incorporated in the training set that we use as an ``unseen" testbed.}
\label{tab:language_groups}
\end{table*}

% \clearpage
\begin{table*}[!hbp]
  \centering
  \label{tabel:dataset_info_seen}
  \resizebox{1\linewidth}{!}{
  \begin{tabular}{llll|llll}
    \toprule
        \textbf{Code} & \textbf{Name} & \textbf{Script} & \textbf{Genus} & \textbf{Code} & \textbf{Name} & \textbf{Script} & \textbf{Genus} \\
        \midrule
        ar-SA & Arabic & Arab & Semitic & is-IS & Icelandic & Latn & Germanic \\
        bn-BD & Bengali & Beng & Indic & ka-GE & Georgian & Geor & Kartvelian \\
        el-GR & Greek & Grek & Greek & km-KH & Khmer & Khmr & Khmer \\
        en-US & English & Latn & Germanic & lv-LV & Latvian & Latn & Baltic \\
        es-ES & Spanish & Latn & Romance & ml-IN & Malayalam & Mlym & Southern Dravidian \\
        fa-IR & Persian & Arab & Iranian & nb-NO & Norwegian & Latn & Germanic \\
        fr-FR & French & Latn & Romance & ro-RO & Romanian & Latn & Romance \\
        he-IL & Hebrew & Hebr & Semitic & sl-SI & Slovenian & Latn & Slavic \\
        hu-HU & Hungarian & Latn & Ugric & ur-PK & Urdu & Arab & Indic \\
        hy-AM & Armenian & Armn & Armenian & zh-CN & Mandarin & Hans & Chinese \\
        id-ID & Indonesian & Latn & Malayo-Sumbawan & zh-TW & Mandarin & Hant & Chinese \\
    \bottomrule
  \end{tabular}}
  \caption{Statistics and description of the dataset used \cite{xu-etal-2022-massive}. The dataset used is a subset of the MASSIVE dataset, selecting 25 different seen languages.\\ \\}
% \end{table*}

% \begin{table*}[ht!]
  \centering
  \label{tabel:dataset_info_unseen}
  \resizebox{1\linewidth}{!}{
  \begin{tabular}{llll|llll}
    \toprule
        \textbf{Code} & \textbf{Name} & \textbf{Script} & \textbf{Genus} & \textbf{Code} & \textbf{Name} & \textbf{Script} & \textbf{Genus} \\
        \midrule
        af-ZA & Afrikaans & Latn & Germanic & my-MM & Burmese & Mymr & Burmese-Lolo \\
        am-ET & Amharic & Ethi & Semitic & nl-NL & Dutch & Latn & Germanic \\
        az-AZ & Azerbaijani & Latn & Turkic & pl-PL & Polish & Latn & Slavic \\
        cy-GB & Welsh & Latn & Celtic & pt-PT & Portuguese & Latn & Romance \\
        da-DK & Danish & Latn & Germanic & ru-RU & Russian & Cyrl & Slavic \\
        de-DE & German & Latn & Germanic & sq-AL & Albanian & Latn & Albanian \\
        fi-FI & Finnish & Latn & Finnic & sv-SE & Swedish & Latn & Germanic \\
        hi-IN & Hindi & Deva & Indic & sw-KE & Swahili & Latn & Bantoid \\
        ja-JP & Japanese & Jpan & Japanese & ta-IN & Tamil & Taml & Southern Dravidian \\
        kn-IN & Kannada & Knda & Southern Dravidian & te-IN & Telugu & Telu & South-Central Dravidian \\
        ko-KR & Korean & Kore & Korean & th-TH & Thai & Thai & Kam-Tai \\
        mn-MN & Mongolian & Cyrl & Mongolic & vi-VN & Vietnamese & Latn & Viet-Muong \\
        ms-MY & Malay & Latn & Malayo-Sumbawan & & & & \\
    \bottomrule
  \end{tabular}}
  \caption{Statistics and description of the dataset used \cite{xu-etal-2022-massive}. The dataset used is a subset of the MASSIVE dataset, selecting 27 different unseen languages.}
\end{table*}

% \lipsum[1-9]
\clearpage
\section{Language Family Experiment}
\label{sec:lang_fam_exp}

Tables \ref{tab:lang_family_performance_bert} and \ref{tab:lang_family_performance_xlmr} provide a comprehensive analysis of language family performance across different training groups. These tables compare the accuracy percentages of the Multilingual BERT and XLM-RoBERTa models, respectively. The results displayed in the tables elucidate the models' capabilities in generalizing from the training data to unseen languages. A clear trend that can be observed is the improvement in performance as the training groups progress from 1 to 4, which suggests that the models benefit from exposure to a wider variety of language families during training. The 'Average' row at the bottom of each table indicates the mean accuracy across all language families, providing an insight into the overall performance enhancement achieved by each model with incremental training diversity.

\begin{table*}[hb!]
\small
\centering
\begin{tabular}{lcccc}
\toprule
\textbf{Language Family} & \textbf{Train Group 1} & \textbf{Train Group 2} & \textbf{Train Group 3} & \textbf{Train Group 4} \\
\midrule
Afro-Asiatic & 52.82\% & 52.93\% & 61.26\% & 61.00\% \\
Atlantic-Congo & 65.71\% & 68.08\% & 70.62\% & 71.79\% \\
Austroasiatic & 64.77\% & 66.78\% & 69.72\% & 70.16\% \\
Austronesian & 66.88\% & 68.66\% & 72.06\% & 72.19\% \\
Dravidian & 64.74\% & 67.97\% & 70.93\% & 71.41\% \\
Indo-European & 67.50\% & 68.61\% & 72.53\% & 72.95\% \\
Japonic & 72.11\% & 71.98\% & 75.80\% & 75.67\% \\
Kartvelian & 68.91\% & 68.89\% & 72.46\% & 72.32\% \\
Koreanic & 64.80\% & 66.46\% & 70.04\% & 69.91\% \\
Mongolic-Khitan & 63.11\% & 66.44\% & 69.71\% & 69.59\% \\
Sino-Tibetan & 62.65\% & 66.29\% & 68.79\% & 70.33\% \\
Tai-Kadai & 63.52\% & 67.89\% & 70.23\% & 71.34\% \\
Turkic & 54.69\% & 56.91\% & 63.54\% & 64.05\% \\
Uralic & 71.49\% & 71.27\% & 75.33\% & 75.15\% \\
\midrule
\textbf{Average} & 65.54\% & 67.07\% & 71.04\% & 71.43\% \\
\bottomrule
\end{tabular}
\caption{Multilingual BERT Performance of Language Families Across Training Groups}
\label{tab:lang_family_performance_bert}
\end{table*}

\begin{table*}[hb!]
\small
\centering
\begin{tabular}{lcccc}
\toprule
\textbf{Language Family} & \textbf{Train Group 1} & \textbf{Train Group 2} & \textbf{Train Group 3} & \textbf{Train Group 4} \\
\midrule
Afro-Asiatic & 75.74\% & 76.23\% & 85.56\% & 85.39\% \\
Atlantic-Congo & 70.86\% & 72.38\% & 83.24\% & 82.73\% \\
Austroasiatic & 74.85\% & 76.04\% & 83.91\% & 83.59\% \\
Austronesian & 78.94\% & 79.83\% & 84.77\% & 84.69\% \\
Dravidian & 81.49\% & 82.20\% & 85.41\% & 85.43\% \\
Indo-European & 80.31\% & 81.21\% & 83.26\% & 83.47\% \\
Japonic & 80.21\% & 81.36\% & 82.67\% & 83.15\% \\
Kartvelian & 80.40\% & 81.53\% & 82.79\% & 83.27\% \\
Koreanic & 79.74\% & 80.91\% & 82.14\% & 82.61\% \\
Mongolic-Khitan & 79.54\% & 81.00\% & 82.20\% & 82.65\% \\
Sino-Tibetan & 79.25\% & 81.00\% & 82.14\% & 82.58\% \\
Tai-Kadai & 79.08\% & 80.81\% & 81.90\% & 82.35\% \\
Turkic & 79.20\% & 80.90\% & 81.96\% & 82.39\% \\
Uralic & 79.24\% & 80.91\% & 81.92\% & 82.47\% \\
\midrule
\textbf{Average} & 79.45\% & 80.48\% & 83.44\% & 83.62\% \\
\bottomrule
\end{tabular}
\caption{XLM-RoBERTa Performance of Language Families Across Training Groups}
\label{tab:lang_family_performance_xlmr}
\end{table*}

\clearpage
\section{Appendix: Language Identification (LID) Experiments}
\label{sec:lid_experiments}

This section presents the results of comprehensive language identification experiments performed across a variety of popular language detection models. The evaluation is detailed in two distinct tables:

Table~\ref{tab:lid-complete} displays the performance of traditional language identification models such as LID-Fasttext, CLD3, CLD2, langid, and LangDetect across multiple languages within the MASSIVE dataset. These results illustrate the effectiveness of each model in correctly identifying the language of given text samples.

Table~\ref{tab:mlm-complete} focuses on the accuracy of multilingual language models, specifically XLM-R and mBERT, alongside adaptations using the MAD-X framework with embeddings from FastText and CLD3. This evaluation aims to show how these advanced models perform in the task of language identification, especially in comparison to more specialized LID tools.

\begin{table*}[ht]
\centering
\resizebox{0.6\linewidth}{!}{
    \begin{tabular}{l|c|c|c|c|c}
    \toprule
    \textbf{Language} & \textbf{LID-Fasttext} & \textbf{CLD3} & \textbf{CLD2} & \textbf{langid} & \textbf{LangDetect} \\
    \midrule
    ar-SA & 94.25 & 86.45 & 81.58 & 91.78 & 94.13 \\
    bn-BD & 99.72 & 97.52 & 89.57 & 96.93 & 99.76 \\
    de-DE & 97.70 & 88.59 & 89.73 & 92.83 & 82.54 \\
    el-GR & 99.68 & 96.91 & 99.77 & 99.84 & 99.64 \\
    en-US & 98.61 & 79.44 & 93.43 & 93.96 & 87.82 \\
    es-ES & 96.20 & 78.24 & 73.14 & 86.87 & 86.55 \\
    fi-FI & 97.70 & 92.91 & 92.90 & 92.08 & 96.09 \\
    fr-FR & 98.35 & 87.53 & 85.23 & 94.77 & 94.80 \\
    hi-IN & 98.44 & 88.21 & 97.83 & 87.94 & 93.54 \\
    hu-HU & 98.54 & 92.24 & 93.89 & 95.34 & 96.71 \\
    hy-AM & 99.90 & 98.37 & 99.92 & 99.17 & 0.00 \\
    id-ID & 87.20 & 65.86 & 73.54 & 72.68 & 89.32 \\
    is-IS & 89.93 & 92.64 & 90.88 & 92.97 & 0.00 \\
    ja-JP & 99.41 & 96.63 & 99.04 & 99.11 & 96.23 \\
    jv-ID & 24.75 & 68.10 & 0.00 & 22.04 & 0.00 \\
    ka-GE & 99.56 & 98.49 & 99.95 & 99.65 & 0.00 \\
    ko-KR & 99.50 & 98.47 & 99.03 & 99.96 & 99.36 \\
    lv-LV & 90.73 & 90.06 & 95.25 & 94.33 & 97.32 \\
    my-MM & 99.93 & 96.90 & 99.97 & 0.00 & 0.00 \\
    pt-PT & 92.17 & 83.42 & 77.39 & 77.74 & 84.05 \\
    ru-RU & 99.27 & 84.48 & 82.35 & 83.79 & 91.32 \\
    vi-VN & 98.41 & 95.85 & 97.26 & 98.62 & 99.53 \\
    zh-CN & 97.55 & 98.07 & 84.33 & 99.64 & 0.00 \\
    zh-TW & 95.76 & 94.19 & 0.03 & 99.31 & 0.00 \\ \midrule
    \textbf{Average} & \textbf{93.89} & \textbf{89.57} & \textbf{83.17} & \textbf{86.31} & \textbf{66.20} \\
    \bottomrule
    \end{tabular}
}
\caption{Per language results of language identification evaluation in MASSIVE.}
\label{tab:lid-complete}
\end{table*}

\begin{table*}[ht]
\centering
\resizebox{0.6\linewidth}{!}{
    \begin{tabular}{l|c|c|c|c|c}
    \toprule
    \textbf{Language} & \textbf{XLMR} & \textbf{mBERT} & \textbf{MAD-X} & \textbf{\begin{tabular}[c]{@{}c@{}}MAD-X\\w/ FastText\end{tabular}} & \textbf{\begin{tabular}[c]{@{}c@{}}MAD-X\\w/ CLD3\end{tabular}} \\
    \midrule
    ar-SA & 79.32 & 78.35 & 75.72 & 71.92 & 67.79 \\
    bn-BD & 83.25 & 80.23 & 78.61 & 76.36 & 74.95 \\
    de-DE & 85.54 & 83.59 & 81.81 & 79.49 & 76.90 \\
    el-GR & 85.07 & 81.74 & 80.93 & 79.56 & 78.51 \\
    en-US & 88.16 & 86.45 & 85.78 & 83.89 & 83.15 \\
    es-ES & 86.18 & 84.97 & 82.58 & 80.97 & 76.43 \\
    fi-FI & 85.24 & 82.55 & 82.55 & 79.86 & 77.07 \\
    fr-FR & 86.48 & 86.11 & 83.69 & 82.35 & 80.03 \\
    hi-IN & 84.63 & 82.38 & 80.73 & 78.14 & 72.73 \\
    hu-HU & 85.68 & 82.65 & 81.57 & 80.13 & 76.40 \\
    hy-AM & 84.23 & 81.20 & 80.43 & 78.78 & 77.91 \\
    id-ID & 86.52 & 84.67 & 82.01 & 76.03 & 69.30 \\
    is-IS & 84.16 & 82.21 & 80.40 & 71.49 & 73.57 \\
    ja-JP & 85.78 & 84.70 & 83.22 & 82.04 & 81.27 \\
    jv-ID & 81.20 & 81.57 & 78.58 & 45.70 & 59.68 \\
    ka-GE & 79.19 & 75.25 & 73.23 & 70.85 & 70.17 \\
    ko-KR & 85.51 & 84.30 & 82.99 & 81.14 & 80.56 \\
    lv-LV & 84.73 & 82.18 & 82.08 & 74.58 & 74.95 \\
    my-MM & 82.18 & 78.01 & 78.48 & 76.36 & 74.98 \\
    pt-PT & 86.35 & 85.27 & 83.59 & 80.56 & 77.77 \\
    ru-RU & 86.65 & 83.96 & 83.52 & 81.74 & 75.45 \\
    vi-VN & 86.48 & 83.32 & 82.52 & 79.72 & 78.61 \\
    zh-CN & 85.41 & 85.24 & 84.23 & 53.09 & 52.69 \\
    zh-TW & 83.73 & 82.55 & 81.27 & 52.79 & 52.45 \\ \midrule
    \textbf{Average} & \textbf{84.65} & \textbf{82.64} & \textbf{81.27} & \textbf{74.90} & \textbf{73.47} \\
    \bottomrule
    \end{tabular}
}
\caption{Per language accuracy score of multilingual language models in MASSIVE.}
\label{tab:mlm-complete}
\end{table*}

% \clearpage
% \section{UNKs Percentage in the Dataset}
% \label{sec:unks_percentage}

% \begin{table}[h!]
%     \centering
%     \begin{tabular}{lll}
%     \toprule
%     Dataset      & Language & UNK Percentage \\
%     \midrule
%     \multicolumn{3}{c}{MASSIVE dataset} \\
%     \midrule
%                  & am-ET    & 6.79\%         \\
%                  & km-KH    & 3.81\%         \\
%                  & vi-VN    & 0.35\%         \\
%     \multicolumn{2}{l}{The rest of the languages} & <1\%           \\
%     \midrule
%     \multicolumn{3}{c}{SemRel dataset} \\
%     \midrule
%                  & amh      & 3.43\%         \\
%                  & hau      & 0.60\%         \\
%                  & ary      & 0.44\%         \\
%     \multicolumn{2}{l}{The rest of the languages} & <0.4\%         \\
%     \midrule
%     \multicolumn{3}{c}{MasakhaNews dataset} \\
%     \midrule
%                  & pcm      & 0.43\%         \\
%                  & eng      & 0.16\%         \\
%     \bottomrule
%     \end{tabular}
%     \caption{UNK percentages in different datasets.}
%     \label{tab:unk_percentages}
% \end{table}

\end{document}